\documentclass[lettersize,journal]{IEEEtran}
\usepackage{amsmath,amsfonts}
\usepackage{algorithm}
\usepackage{algpseudocode}
\usepackage{array}
\usepackage[caption=false,font=normalsize,labelfont=sf,textfont=sf]{subfig}
\usepackage{textcomp}
\usepackage{stfloats}
\usepackage{url}
\usepackage{verbatim}
\usepackage{graphicx}
\usepackage{cite}
\usepackage{svg}
\usepackage{booktabs}
\usepackage{multirow}
\usepackage{makecell}

\begin{document}

\title{
Tri-Band Channel Measurement-Enabled Multi-Layer Digital Twin for Terahertz Wireless Data Centers}

\author{Mingjie Zhu, Ziming Yu, Guangjian Wang and Chong Han,~\IEEEmembership{Senior~Member,~IEEE}
\thanks{
Mingjie~Zhu is with the Terahertz Wireless Communications (TWC) Interdisciplinary Research Center, Shanghai Jiao Tong University, Shanghai 200240, China (E-mail: mingjie.zhu@sjtu.edu.cn). 

Ziming Yu and Guangjian Wang are with Huawei Technologies Company Ltd., Chengdu 611731, China (e-mail: yuziming@huawei.com).

Chong~Han is with the Terahertz Wireless Communications (TWC) Interdisciplinary Research Center, and Cooperative Medianet Innovation Center (CMIC), School of Information Science and Electronic Engineering, Shanghai Jiao Tong University, Shanghai 200240, China (E-mail:~chong.han@sjtu.edu.cn). 
}
}



\maketitle
\begin{abstract}
The rapid growth of AI computing has driven increasing demands for flexible and high-capacity data-center interconnections. Owing to its ultra-wide bandwidth and high spatial reuse capability, terahertz (THz) communication has emerged as a promising solution for future wireless data centers, while digital twins (DTs) enable efficient wireless planning and real-time optimization. In this work, a measurement-driven multi-layer DT framework is proposed for THz wireless data centers, where the physical twin at the measurement layer, channel twin at the construction layer, system performance at the evaluation layer, and resource allocation from the manipulation layer are progressively constructed from bottom to top.
First, extensive channel measurements are conducted at 140, 220, and 300~GHz to characterize frequency-dependent propagation behaviors. Based on the tri-band measurements, a measurement-calibrated physical twin is established by jointly optimizing the geometry, material, antenna, and hybrid propagation models. On top of the physical twin, a line-of-sight (LoS)-aware implicit neural field is developed to construct an AI channel twin for efficient channel reconstruction. The proposed AI twin learns location-dependent channel statistics from the calibrated twin, enabling real-time prediction of received power and LoS probability. Moving further up and building upon the reconstructed channel field, a system-level evaluation layer is then derived to analyze coverage and interference for both AP-to-rack and rack-to-rack communications.
Experimental results show that the proposed AI twin achieves lower power reconstruction error than existing neural-field baselines while maintaining real-time inference capability. Moreover, the ceiling-mounted AP deployment achieves over 90\% coverage under a 10~dB signal-to-interference-plus-noise ratio (SINR) threshold, demonstrating the effectiveness of the proposed DT framework for THz wireless data-center planning and optimization.
\end{abstract}




\section{Introduction}
Modern data centers play a critical role in supporting cloud computing, artificial intelligence (AI), and large-scale distributed storage systems~\cite{11605106, 7393451}.
With the rapid growth of data-intensive applications, information traffic within a data center has increased dramatically, calling for a more flexible and low-latency architecture~\cite{8367741, 10.1145/1879141.1879175}.
To this end, Terahertz (THz) wireless data center (WDC) has emerged as a promising connection architecture for next-generation AI clusters~\cite{11605106}, due to its ultra-wide bandwidth and highly directional transmission~\cite{9766110, 9665432, 6005345}.
These characteristics make THz wireless links capable of supporting extremely high data rates with low latency and high energy efficiency, while also enabling dynamic and reconfigurable network topologies.
However, the practical deployment and operation of THz wireless data centers require reliable network planning, interference management, and real-time resource allocation. Such functions cannot rely solely on static models or offline simulations, motivating the need for an accurate and adaptive digital twin (DT)~\cite{9120192, 11261676, 10742102, 10234421} that can faithfully represent the wireless environment and support system-level decision-making.



The propagation mechanism in a wireless data center is fundamental for THz-WDC optimization and deployment.
THz propagation is highly sensitive to environmental dynamics, which are prevalent in realistic data-center scenarios. As a result, reliable network planning, interference management, and real-time resource allocation rely on an accurate and adaptive digital twin (DT)~\cite{9120192,11261676,10742102,10234421} that can accurately perceive the physical environment, faithfully reconstruct the wireless propagation characteristics, efficiently evaluate network performance, and ultimately support intelligent network optimization. Such a closed-loop DT enables continuous interactions between the physical and virtual worlds, providing a foundation for autonomous wireless network management.
However, realizing such a DT remains challenging.
First, the DT must be grounded in realistic propagation measurements in order to capture the frequency-dependent characteristics of THz channels.
Second, the DT requires a channel representation that is both physically consistent and measurement validated, so that the virtual model faithfully reproduces the real propagation behaviors.
Third, the channel representation should support real-time interaction, requiring substantially lower computational complexity than conventional RT simulations.
Finally, the reconstructed channel knowledge should be translated into network-level intelligence, enabling coverage evaluation, deployment planning, and resource optimization.

Since the fidelity of a wireless DT fundamentally relies on accurate channel knowledge, obtaining realistic propagation data becomes the first and foremost prerequisite for DT construction. Therefore, channel measurement serves as the foundation for DT construction. A growing body of literature has investigated channel measurements for THz communications~\cite{9794668,9039668,10001052,10609428}. Existing measurement campaigns have characterized path loss, delay dispersion, angular characteristics, and blockage effects in various THz scenarios~\cite{10346993,11568954}. In~\cite{10001052}, channel measurement is carried out at 140~GHz in a wireless data center, whereas physical-layer evaluation is not discussed. A cluster-based model is established for THz channel representation in a data center in~\cite{9039668}, and a hybrid model for THz wireless data centers is utilized in~\cite{10609428}. However, measurements in wireless data-center environments remain relatively limited, particularly for multi-band characterization and DT-oriented applications.

To faithfully reproduce the measured propagation behaviors, an effective DT further requires a channel representation that is both physically consistent and measurement validated. Built upon measurement data, the channel twin of wireless DTs has been predominantly realized through ray-tracing (RT)-based methodologies~\cite{10643616,11027522,10855530}. By leveraging site-specific geometries and electromagnetic propagation mechanisms, RT enables the reconstruction of multipath propagation characteristics with high physical interpretability. Recent studies have further incorporated calibration techniques to improve the consistency between simulations and measurements~\cite{10643616}, while experimental validations have demonstrated certain feasibility of RT for THz channel characterization~\cite{10855530}. 

Besides reconstruction fidelity, a practical DT should also support real-time interaction with the physical world. To this end, AI-enabled channel twins have recently emerged as a promising solution for wireless DT construction~\cite{10234421,han2026aimeetsterahertzsurvey,8985539,11319342,hu2024transfer,11316498}. By learning the underlying mapping between environmental features and channel responses, AI-based methods can substantially reduce computational overhead while maintaining satisfactory prediction accuracy. Existing studies have explored generative models, diffusion models, generative adversarial networks (GANs), and transformer-based architectures for channel generation and channel estimation. In particular, the integration of physics-based knowledge and data-driven learning has demonstrated significant potential for accelerating DT construction in future 6G networks~\cite{10234421}. However, most existing AI channel twins focus on generic communication scenarios and often overlook the blockage-sensitive propagation characteristics and highly structured layouts encountered in THz wireless data centers. Furthermore, the relationship between AI-based channel twins and measurement-calibrated physical models remains insufficiently investigated.

Once an accurate channel twin has been established, it can further support network-level decision making. DTs have been increasingly employed for network-level optimization and autonomous decision making~\cite{11202728,10742102}. By coupling environment-aware channel prediction with optimization algorithms, DTs enable efficient deployment planning, coverage analysis, interference management, and resource allocation. Nevertheless, existing optimization-oriented DTs generally rely on either computationally intensive RT simulations or simplified channel abstractions, limiting their ability to provide both high fidelity and real-time responsiveness in complex indoor environments.

\begin{figure}
	\centering
	\includegraphics[width=1\linewidth]{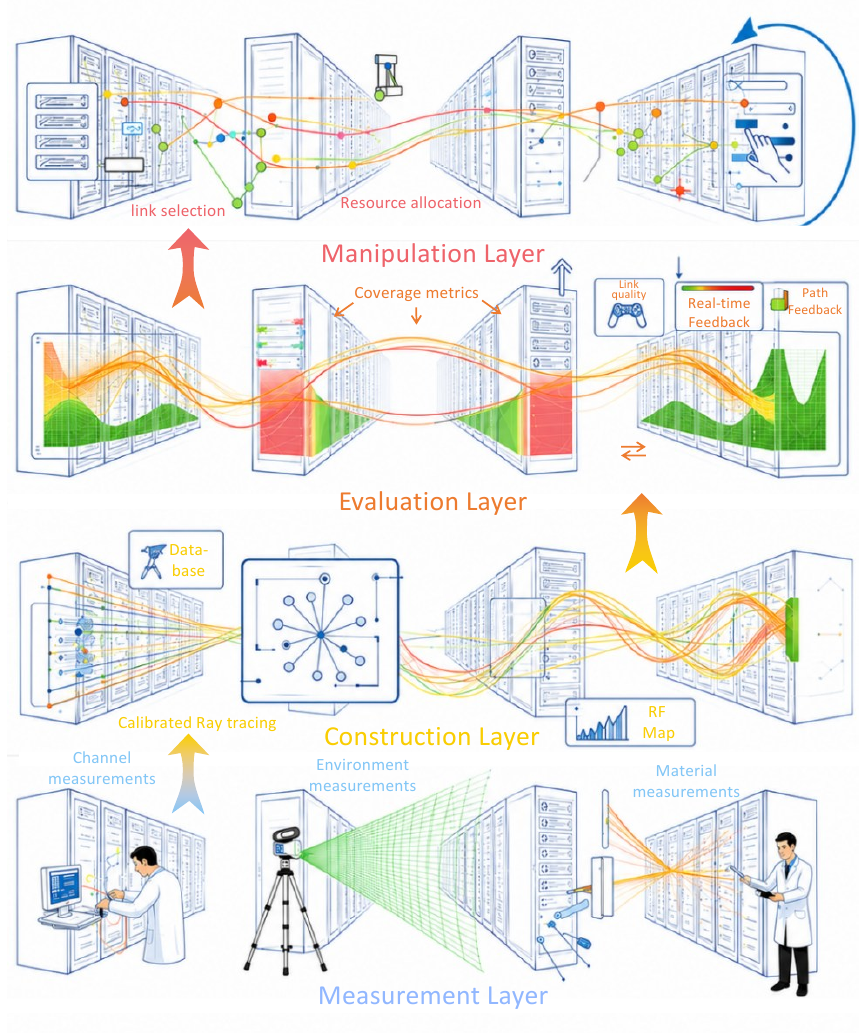}
	\caption{Multi-layer DT architecture.}
	\label{fig:dta}
\end{figure}

Motivated by the above observations, we propose a measurement-driven multi-layer digital twin for THz wireless data centers, as illustrated in Fig.~\ref{fig:dta}. The proposed framework is organized into four functional layers from bottom to top. The measurement layer constitutes the foundation of the physical twin, aiming to capture the physical characteristics of the wireless environment through multi-source measurements. It includes the channel measurement, environment measurement, and material measurement.
The construction layer builds a channel map of a wireless data center, including a measurement-calibrated RT twin and an AI twin. The calibrated RT twin mitigates the discrepancies between RT and measurements, while the AI twin accelerates the channel representation via AI channel models. The channel twins construct a virtual counterpart capable of reproducing the observed propagation behaviors.
The evaluation layer translates the reconstructed channel statistics into system metrics such as coverage, interference, and achievable rate, bridging the gap between channel-level predictions and network-level performance assessment.
Finally, the manipulation layer exploits the evaluation results to perform access-point deployment, beam management, user association, and interference-aware optimization, thereby enabling closed-loop planning and real-time decision making.
By jointly leveraging multi-band channel measurements, physics-based modeling, and AI-assisted channel reconstruction, the multi-layer DT achieves both high fidelity and computational efficiency, providing a practical platform for channel characterization, network evaluation, and intelligent optimization in future THz wireless data centers.

The main contributions of this paper are summarized as follows:

\begin{itemize}
    \item We propose a measurement-driven multi-layer DT framework for THz wireless data centers. The proposed framework integrates a geometry twin constructed from LiDAR point clouds, a material twin calibrated using THz-TDS measurements, and a measurement-calibrated channel twin combining ray tracing and channel parameter extraction. By jointly incorporating geometric, material, and propagation information, the multi-layer DT significantly improves the fidelity and physical consistency of site-specific THz DTs.

	\item For the measurement layer, we conduct a comprehensive tri-band channel measurement campaign in a realistic wireless data-center environment at 140, 220, and 300~GHz. Extensive channel characteristics, including path loss, delay dispersion, angular properties, blockage effects, and multipath clustering behaviors, are extracted and analyzed, providing valuable insights into frequency-dependent THz propagation mechanisms in data-center scenarios.

	\item For the construction layer, we develop a blockage-aware AI channel twin based on an improved implicit neural field (INF) to enable real-time channel reconstruction and system evaluation. Different from conventional coordinate-based neural fields, the proposed model incorporates distance-aware propagation priors, grid-based LoS probability maps, and a LoS-adaptive dual-expert architecture to explicitly capture the distinct propagation behaviors of LoS and NLoS links in blockage-dominated data-center environments. The proposed AI twin achieves high channel reconstruction accuracy while substantially reducing the computational complexity compared with full RT simulations.
	
	\item For the evaluation layer, we derive analytical coverage and rate evaluation models for THz wireless data centers. The resulting closed-form coverage expressions establish a direct connection between channel prediction and network-level performance metrics, enabling efficient coverage estimation, interference analysis, and deployment evaluation. The proposed framework provides a complete closed-loop workflow from channel measurement and DT construction to intelligent network optimization.
	
\end{itemize}

The remainder of the paper is organized from bottom to top of the multi-layer DT. The measurement layer, construction layer, and evaluation layer are introduced respectively in Sections~\ref{section: physical}, section~\ref{section: channel}, and section~\ref{section: evaluation}. Finally, Section~\ref{section: con} draws the conclusion

\section{Measurement layer}
\label{section: physical}



The physical characteristics of the WDC are constituted through multi-source measurements, including channel measurement, environment measurement, and material measurement. This section presents the measurement campaign at 140, 220, and 300 GHz and the extracted propagation characteristics, which serve as the physical foundation of the multi-layer DT. 

\subsection{Tri-band channel measurement}
\begin{figure}
	\centering
	\subfloat[]{
		\includegraphics[width=0.8\linewidth]{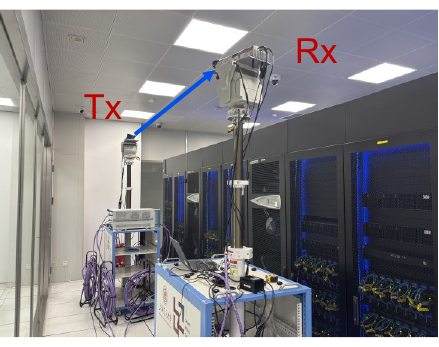}
	}
	\hfill
	\subfloat[]{
		\includegraphics[width=0.85\linewidth]{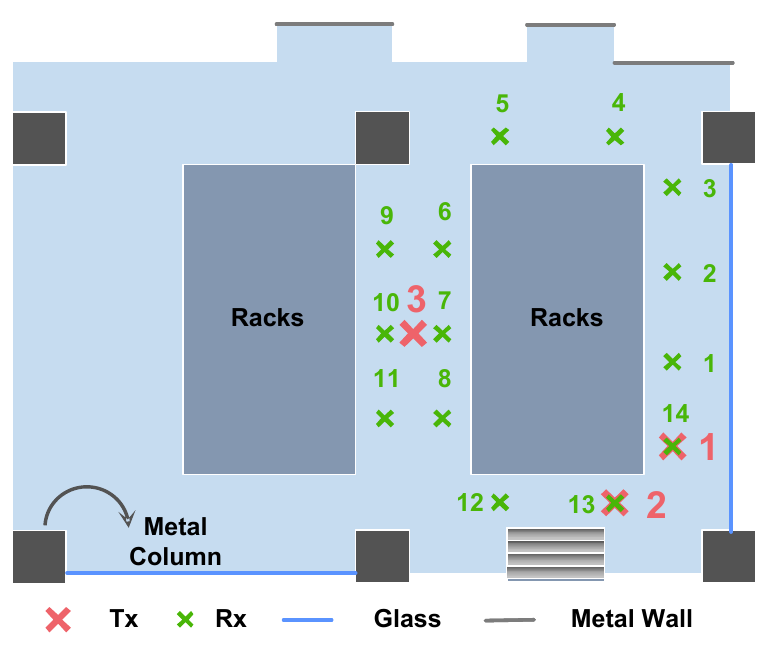}
	}
	\caption{THz channel measurement in the data center. (a) A photo of our THz sounder system; (b) Measurement topology in the data center scenario.}
	\label{fig:scheme}
\end{figure}

\subsubsection{Measurement Setups}
Fig.~\ref{fig:scheme}(a) shows the channel measurement setup deployed inside the wireless data-center environment. The transmitter (Tx) location, receiver (Rx) locations, antenna configurations, and scanning procedures are maintained consistently across all frequency bands to ensure fair cross-band comparisons.
For the 140~GHz and 220~GHz bands, a correlation-based channel measurement system is employed, providing an effective measurement bandwidth of 1.536~GHz. For the 300~GHz band, a VNA-based channel sounder is adopted to capture the channel frequency responses (CFRs) over the 290--310~GHz frequency range. Different measurement systems are used because the two platforms provide complementary frequency coverage. The significantly larger bandwidth of 20~GHz provides a delay resolution of 0.05~ns, corresponding to a distance resolution of approximately 1.5~cm.
During the measurements, the Tx remains fixed while a directional scanning scheme is implemented at the Rx. Specifically, the azimuth angle is scanned from $0^{\circ}$ to $355^{\circ}$ with a step size of $5^{\circ}$, while the zenith angle ranges from $-20^{\circ}$ to $20^{\circ}$ with the same angular resolution. This configuration enables the acquisition of high-resolution angular and delay-domain channel characteristics across all three frequency bands.
Prior to each measurement campaign, back-to-back calibration is performed to eliminate the responses of cables, mixers, amplifiers, and other hardware components, thereby ensuring that the extracted channel characteristics accurately represent the propagation behavior of the wireless data-center environment.
The measurement configurations are summarized in Table~\ref{tab:cfg}.


\begin{table*}[t]
\caption{Measurement Configuration.}
\label{tab:cfg}
\centering
\small
\setlength{\tabcolsep}{6pt}
\renewcommand{\arraystretch}{0.9}

\begin{tabular}{lcc}
\toprule
Parameter & 300 GHz (VNA) & 140/220 GHz (Correlation-based) \\
\midrule
Center Frequency & 300 GHz & 140/220 GHz \\
Bandwidth & 20 GHz & 1.536 GHz \\
Frequency Points & 2001 & 2048 \\
IF Bandwidth & 1 kHz & 1.536 GHz \\
\midrule
Transmit Power & \multicolumn{2}{c}{10 dBm} \\
Tx/Rx Antenna Type & \multicolumn{2}{c}{Horn antenna} \\
Tx/Rx Height 
& \multicolumn{2}{c}{Tx: 2.4 m (rack-to-rack), 2.7 m (AP); Rx: 2.4 m (LoS), 1.7 m (NLoS)} \\
Angular Scanning
& \multicolumn{2}{c}{Azimuth: $5^\circ$, $0^\circ$--$355^\circ$; Zenith: $5^\circ$, $-20^\circ$--$20^\circ$} \\
Antenna Gain & \multicolumn{2}{c}{26 dBi} \\
Half-Power Beamwidth & \multicolumn{2}{c}{Horizontal: $8^\circ$; Vertical: $6^\circ$} \\
\bottomrule
\end{tabular}
\end{table*}

\subsubsection{Measurement Scenario}
As illustrated in Fig.~\ref{fig:scheme}(b), the measurements cover four cases in the data center, corresponding to line-of-sight (LoS) and non-line-of-sight (NLoS) conditions for both Rack-to-Rack and AP-to-User links. A total of 29 Tx-Rx locations are measured. 
Tx~$1$ is positioned alongside the rack, with 9 LoS and 3 NLoS Rxs alongside the rack, acting as the rack-to-rack communication. Tx~$2$ is positioned in front of the rack, with 5 Rxs (Rx~$4-8$) alongside the rack, representing the rack-to-rack communication in the other direction. Tx~$3$ simulates an access point (AP), located at the center of the ceiling, with 9 LoS and 3 NLoS Rxs alongside the rack, simulating AP-to-rack communication.


\subsection{Channel Parameter Extraction}



The first step of the physical twin is to extract channel parameters from the measured directional channel response. After directional scanning, the channel response can be represented in the delay-angle domain as

\begin{equation}
P(\tau,\theta,\varphi)
=
|h(\tau,\theta,\varphi)|^2,
\end{equation}
which forms the power-angle-delay profile (PADP).
The signal model is
\begin{equation}
     h(\tau, \theta, \varphi)=\sum_{\ell=1}^{L} \alpha_{\ell} \delta(\theta-\theta_{\ell})\delta(\varphi-\varphi_{\ell}) \delta(\tau-\tau_{\ell})
\end{equation}
where $\alpha_{\ell}$ and $\tau_{\ell}$ stand for the complex amplitude and delay of arrival (DoA) at the Rx of the $\ell^{\mathrm{th}}$ path among all the $L$ MCPs.



The MPC parameters are then extracted from the PADP through an iterative peak-search procedure. At each iteration, the strongest local maximum exceeding a predefined threshold is identified as
\begin{align}
    P(\tau) =\left\{\begin{array}{rcl} \mathop{\mathrm{max}}\limits_{\theta_{\ell},~ \varphi_{\ell}}|h(\tau_{\ell},\theta_{\ell}, \varphi_{\ell})|^2, & |h(\tau_{\ell},\theta_{\ell}, \varphi_{\ell})|^2>\mathrm{thr}\\
    0, &\mathrm{otherwise}.
\end{array}\right.
\end{align}


The detected peak provides an initial estimate of the corresponding MPC. Each path is therefore represented by a parameter vector
\begin{equation}
    \mathbf{\Theta}_{\ell}=\{\tau_{\ell},~\theta_{\ell},~\varphi_{\ell},~\alpha_{\ell}\}.
\end{equation}

After extracting the multipath parameters from the measurements, frequency-dependent channel characteristics can be further analyzed and compared across different THz bands.
The cross-band similarity is first quantified through the correlation coefficient of the path loss measurements. For two frequency bands $f_i$ and $f_j$, the correlation coefficient is calculated as

\begin{equation}
\label{eq:correlationmatrix}
	\rho_{f_i.f_j}=\frac{\sum_{n=1}^{N}\left(\mathrm{PL}_{n}^{f_i}-\overline{\mathrm{PL}^{f_i}}\right)\left(\mathrm{PL}_{n}^{f_i}-\overline{\mathrm{PL}^{f_j}}\right)}{\sqrt{\sum_{n=1}^{N}\left(\mathrm{PL}_{n}^{f_j}-\overline{\mathrm{PL}^{f_i}}\right)^2}\sqrt{\sum_{n=1}^{N}\left(\mathrm{PL}_{n}^{f_j}-\overline{\mathrm{PL}^{f_j}}\right)^2}}
\end{equation}

The resulting correlation matrix provides insights into the consistency of propagation characteristics among different frequency bands and reveals the degree of frequency dependency in the data-center environment.







\subsection{Physical Twin Observations}
\begin{figure*}
    \centering
    
    \hspace{-0.1\columnwidth}
    \subfloat[]{
        \includegraphics[width=0.3\linewidth]{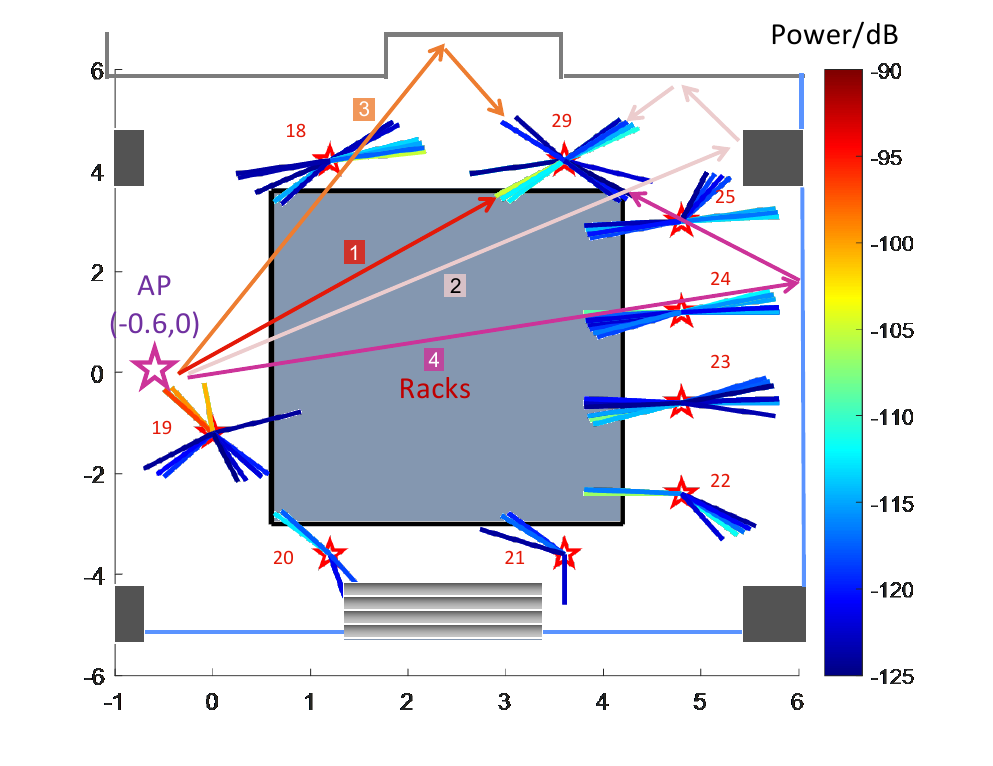}
    }
    \hspace{0.03\columnwidth}
    \subfloat[]{
    	\includegraphics[width=0.27\linewidth]{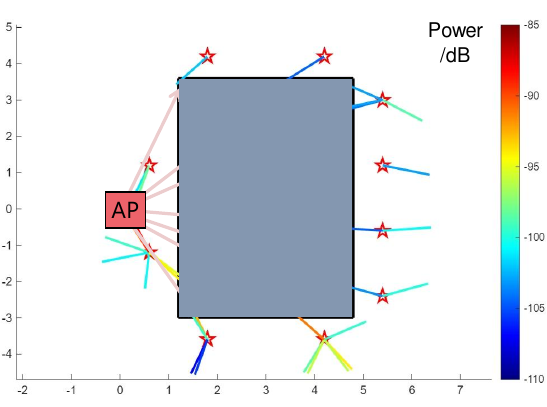}
    }
    \hspace{0.03\columnwidth}
    \subfloat[]{
        \includegraphics[width=0.27\linewidth]{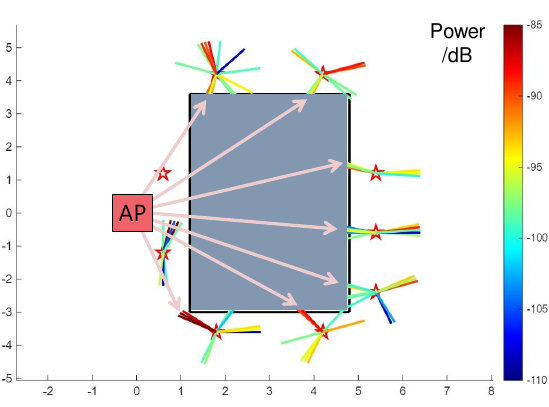}
    }
    \hfill
    \subfloat[]{
    	\includegraphics[width=0.31\linewidth]{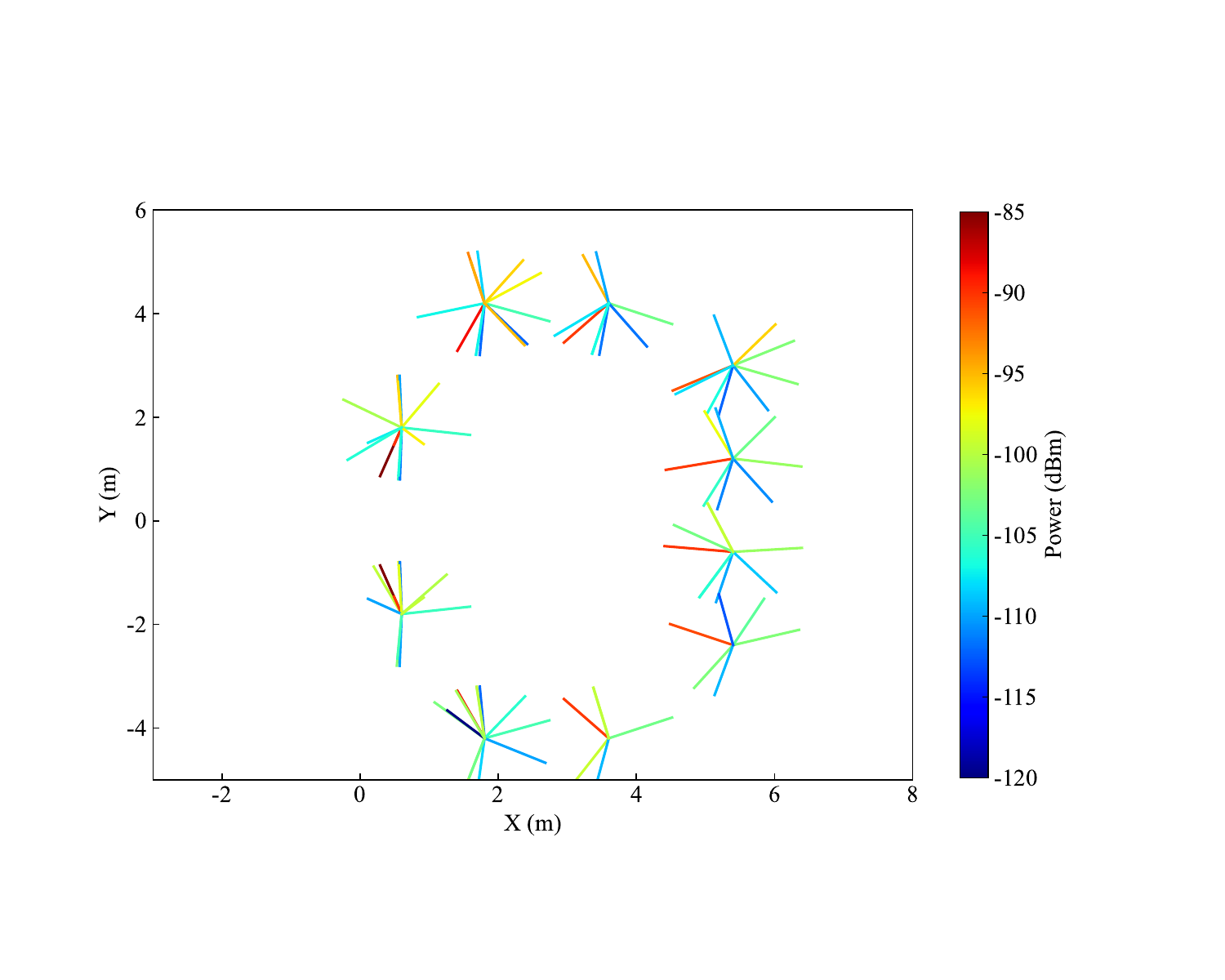}
    }
    \subfloat[]{
    	\includegraphics[width=0.31\linewidth]{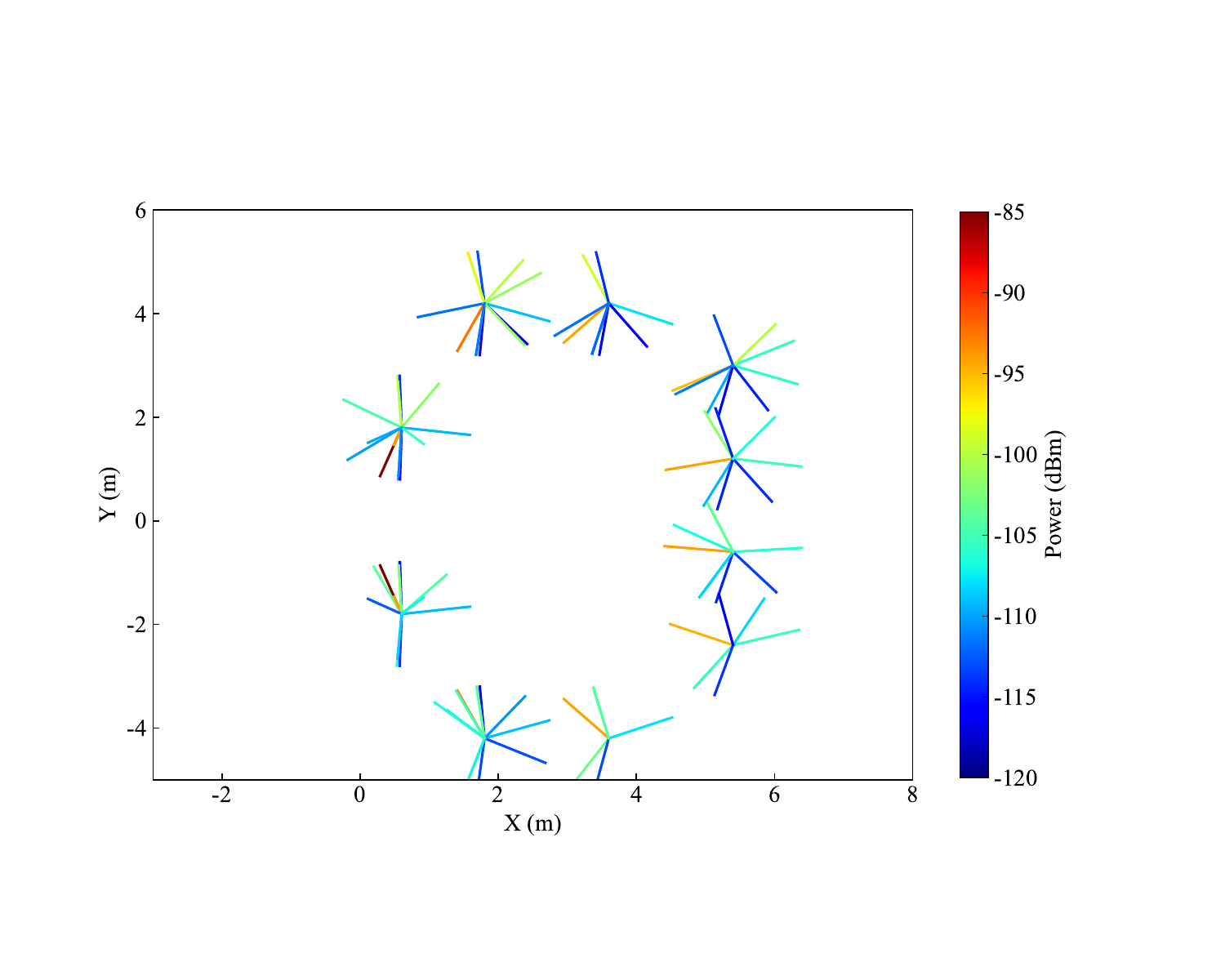}
    }
    \subfloat[]{
    	\includegraphics[width=0.31\linewidth]{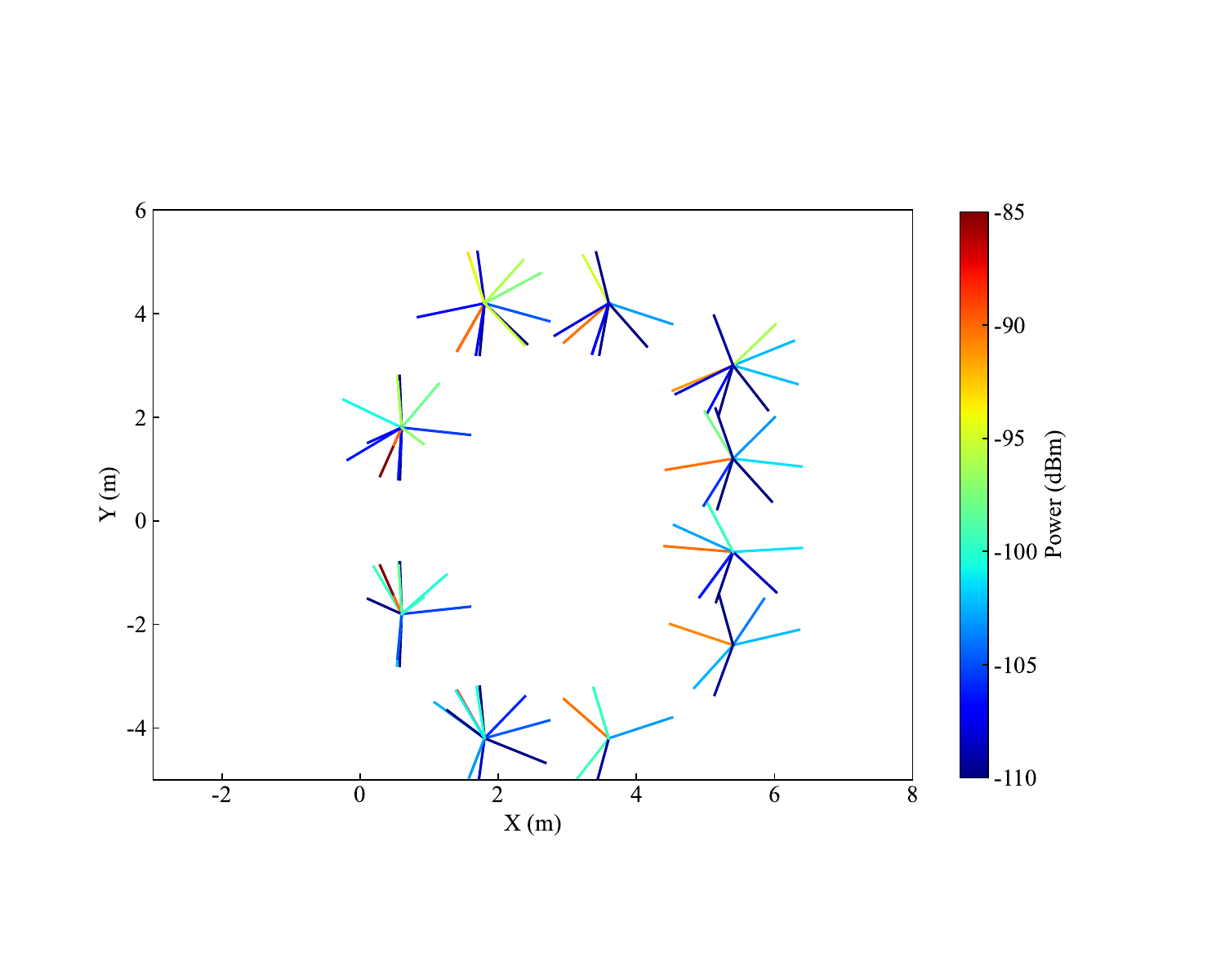}
    }
    \caption{MPC Twins. (a)-(c) Physical twin of 300, 220, and 140 GHz; (d)-(f) RT twin of 300, 220, and 140 GHz.}
    \label{fig:mpcs}
\end{figure*}

The physical twin first provides a comprehensive characterization of the multipath propagation mechanisms inside the wireless data-center environment. The spatial distributions of the extracted MPCs for the three measured frequency bands together is presented in Fig.~\ref{fig:mpcs}(a) to (c). The MPC trajectories are projected onto the horizontal plane to visualize the dominant propagation mechanisms between the transmitter and all receiver locations.

For all three frequency bands, the strongest MPC is consistently dominated by the LoS component, demonstrating that the direct propagation path remains the primary communication mechanism inside the open rack corridors. Besides the LoS path, strong reflected components can be observed from the rear glass wall and metallic rack surfaces, while a large number of weaker diffuse MPCs are distributed around the dominant reflections. These observations indicate that both specular reflections and diffuse scattering contribute significantly to the THz channel characteristics in practical data-center environments. Notably, the LoS path is observed in 52\% of the measured receiver locations, while the remaining 48\% are dominated by NLoS propagation due to rack blockage.
The difference in the number of extracted MPCs is mainly attributed to the different measurement systems.


\begin{table}[t]
	\caption{Correlation Matrix among the three bands.}
	\label{coma}
	\centering
	\begin{tabular}{c|ccc}
		\toprule
		Frequency (GHz)	 & 140 & 220 & 300 \\
		\midrule
		
		140 & 1 & 0.943 & 0.909 \\
		220 & 0.943 & 1 & 0.981 \\
		300 & 0.909 & 0.981 & 1 \\
		
		\bottomrule
	\end{tabular}
\end{table}
Besides verifying the reconstructed propagation paths, it is also important to investigate whether the propagation characteristics remain consistent across different THz frequency bands.
To quantitatively investigate the similarity among different THz frequency bands, Table~\ref{coma} summarizes the correlation coefficients of the measured path losses, as calculated in \eqref{eq:correlationmatrix}.
Strong correlations are observed among all three frequency bands. These results indicate that the spatial variations of the large-scale channel characteristics are highly consistent across the three frequency bands. In particular, the correlation of 0.981 between 220 and 300~GHz suggests that the two higher-frequency bands exhibit highly similar spatial propagation patterns in the investigated data-center environment, while the slightly lower correlation between 140 and 300~GHz indicates a relatively stronger frequency-dependent variation over a wider frequency separation. Nevertheless, the correlation remains above 0.9 for all frequency pairs, demonstrating substantial cross-band consistency and suggesting that the dominant propagation structures, such as blockage and major reflection regions, are largely shared across the three bands.



\begin{figure}[t]
	\centering
	\includesvg[width=0.95\linewidth]{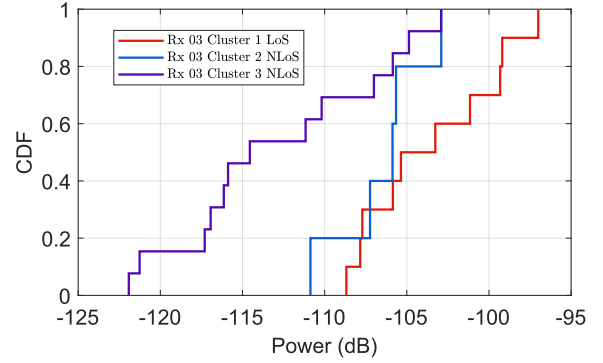}
	\caption{Power distribution of the representative Tx 1--Rx 3 link at 300 GHz.}
	\label{fig:CDF}
\end{figure}

\section{Construction layer}
\label{section: channel}
Based on the tri-band measurements and the extracted channel characteristics, the channel twin is constructed to provide an accurate yet computationally efficient representation of the THz propagation environment. The proposed channel twin consists of two complementary components, i.e., a calibrated RT twin and an AI twin. The former improves the physical fidelity of RT using measurement data, while the latter enables efficient channel reconstruction through a learning-based neural representation.

\subsection{Measurement-calibrated RT Twin}
The calibrated RT twin serves as the physical foundation of the proposed channel twin, providing both an accurate representation of the propagation environment and reliable supervision for the subsequent AI twin. To maximize the consistency between simulations and practical measurements, the RT model is calibrated using the measured tri-band channel data from multiple aspects.
Specifically, the discrepancies between RT and measurements are mitigated through four complementary calibrations:
\begin{itemize}
	\item \textbf{Geometry Calibration} To establish a geometry-aware DT of the THz data center environment, a LiDAR-based three-dimensional reconstruction framework was adopted to capture the detailed spatial structure of the measurement scenario. Compared with conventional manually constructed layouts, LiDAR scanning enables a more accurate representation of rack deployment, corridor structures, cable distributions, and other irregular objects that significantly affect THz propagation characteristics.
	
	\item \textbf{Material Calibration} Frequency-dependent electromagnetic material properties are extracted using THz time-domain spectroscopy (THz-TDS), enabling accurate characterization of dielectric constants, conductivity, and reflection behaviors across the 140/220/300 GHz bands~\cite{11153056}.

    \item \textbf{Antenna Calibration} THz communications could reply on high-gain directional horn antennas to achieve narrow-beam transmission, thereby reducing interference and improving spatial reuse. To accurately reproduce the practical propagation characteristics, the measured antenna radiation patterns are incorporated into the RT simulation.
    
	\item \textbf{Propagation Calibration} Deterministic RT alone is insufficient for capturing the diffuse multipath characteristics in blockage-rich THz data-center environments. A hybrid propagation model~\cite{9466322} combining deterministic specular components and diffuse scattering mechanisms is introduced to compensate for unresolved multipath components and rough-surface scattering effects in blockage-rich THz data-center scenarios.

\end{itemize}

Specifically, the calibration process is formulated as the following optimization problem:
\begin{equation}
	\theta^{*} = \mathrm{arg} \min\limits_{\theta} \sum_{n}\mathcal{D}\left(\mathbf{c}_{meas}^{(n)},\mathbf{c}_{\mathrm{RT}}^{(n)}(\theta)\right)
\end{equation}
where $\theta={\theta_g,\theta_m,\theta_g}$ denotes the set of geometry, material, and propagation parameters to be calibrated, respectively. $\mathbf{c}_{meas}^{(n)}$ and $\mathbf{c}_{\mathrm{RT}}^{(n)}$ represent the measured and simulated channel descriptors at the n-th location, including the multipath power, delay, and angular characteristics, i.e.,
\begin{equation}
	\mathbf{c}=\left\{P_l, \tau_l, \varphi_l, \theta_l\right\}_{l=1}^{N}
\end{equation}


At the geometry layer, high-resolution LiDAR point clouds are employed to reconstruct the rack structures, corridors, and blockage boundaries of the data center, thereby reducing geometric approximation errors in the RT environment.
The reconstructed point cloud is further converted into triangular meshes used by the RT engine, i.e.,
\begin{equation}
\mathcal{G}
=
\left\{
\mathbf{v}_i,\,
\mathbf{f}_j
\right\},
\end{equation}
where $\mathbf{v}_i$ denotes the mesh vertices and $\mathbf{f}_j$ represents the corresponding triangular facets. The geometric model directly determines the visibility, reflection paths, and blockage relationships in the RT simulation.

At the material layer, frequency-dependent dielectric parameters are extracted using THz-TDS measurements. The material properties directly affect the Fresnel reflection coefficients in the RT simulation. The complex permittivity of representative data-center materials is characterized through THz-TDS measurements. The frequency-dependent dielectric property is expressed as
\begin{equation}
\tilde{\varepsilon}
=
\varepsilon_r
-
j
\frac{\sigma}
{2\pi f\varepsilon_0},
\end{equation}
where $\varepsilon_r$ denotes the relative permittivity, $\sigma$ is the conductivity, and $\varepsilon_0$ is the vacuum permittivity.
Accordingly, the reflection loss used in the RT calibration is computed as
\begin{equation}
RL
=
-20\log_{10}
|\Gamma|,
\end{equation}
which directly determines the reflected-path attenuation in the RT simulation.

The antenna radiation characteristics are incorporated into the RT simulation through the measured directional antenna patterns. The received power of each propagation path is given by

\begin{equation}
P_r=P_t+G_t(\theta_t,\phi_t)+G_r(\theta_r,\phi_r)-PL,
\end{equation}
where $G_t(\cdot)$ and $G_r(\cdot)$ denote the directional gains of the transmitting and receiving antennas, respectively. The measured antenna patterns are imported into the RT engine to accurately reproduce the beamwidth and side-lobe characteristics observed during channel measurements.


Although the calibrated geometry, material properties, and antenna patterns significantly improve the deterministic propagation accuracy, conventional RT still fails to capture diffuse scattering caused by rough surfaces, cable trays, server equipment, and irregular rack structures.
Therefore, a hybrid propagation model is further introduced to compensate for the missing diffuse multipath components.
Specifically, the channel response is modeled as the superposition of deterministic specular components and diffuse scattering components, as
\begin{equation}
	h_{hybrid}(\tau,\theta,\varphi)=h_{RT}(\tau,\theta,\varphi)+h_{s}(\tau,\theta,\varphi),
	\label{equation: hybrid}
\end{equation}
where
\begin{align}
	h_{s}(\tau,\theta,\varphi)=&\sum_{\ell}^{L}\sum_{p}\alpha_{\ell,p} \delta(\tau - \tau_{\ell,p}) \cdot \delta(\theta - \theta_{\ell,p}) \cdot\delta(\varphi - \varphi_{\ell,p}) \notag\\
	+& \sum_{q}^{L_s}\sum_{s}\alpha_{q,s} \delta(\tau - \tau_{q,s}) \cdot \delta(\theta - \theta_{q,s}) \cdot\delta(\varphi - \varphi_{q,s}),
\end{align}
where $p$ denotes the $p^{th}$ subpath of ray-traced $\ell^{th}$ cluster, while $q$ stands for the $q^{th}$ subpath of non-RT $s^{th}$ cluster. $L$ and $L_s$ represent the number of ray-traced clusters and non-RT clusters, respectively.

\begin{equation}
	\tau_{\mathrm{RMS},\ell}=\sqrt{\frac{\sum_{p}P_{p,\ell}\left(\tau_{p,\ell}-\bar{\tau}_{\ell}\right)^2}{\sum_{p}P_{p,\ell}
		}
	},
	\label{equation:rms_delay}
\end{equation}

\begin{equation}
	f_{\mathrm{meas}}(i)=\frac{P_i^{\mathrm{meas}}}{\sum_{i=1}^{M}P_i^{\mathrm{meas}}},
	\label{equation:pmf_meas}
\end{equation}

Comparing the measured and RT-generated MPC distributions in Fig.~\ref{fig:mpcs}, it can be observed that the calibrated RT twin successfully reproduces the dominant propagation paths and their corresponding arrival directions at all three frequency bands. Nevertheless, the conventional specular-only RT fails to capture numerous low-power diffuse components observed in the measurements, especially around rack edges and rear-wall reflections. These missing MPCs motivate the introduction of the proposed hybrid propagation model presented in Section III-B.

Although the dominant propagation mechanisms can be reproduced by the calibrated RT twin, discrepancies still exist for diffuse scattering components.
To further investigate the discrepancies between the measured and simulated channels, a representative receiver (Rx3 at 300~GHz) is selected for detailed analysis. Fig.~\ref{fig:CDF} illustrates the cumulative power distributions of the three dominant multipath clusters extracted from the measurements. Three well-separated clusters can be identified, corresponding to the LoS component, the rear-wall reflection, and the rack reflection, respectively. Besides the dominant paths, each cluster also contains numerous weak MPCs caused by diffuse scattering and angular spreading, leading to a long-tail power distribution.

\begin{figure}
	\centering
    \subfloat[]{
        \includegraphics[width=0.77\linewidth]{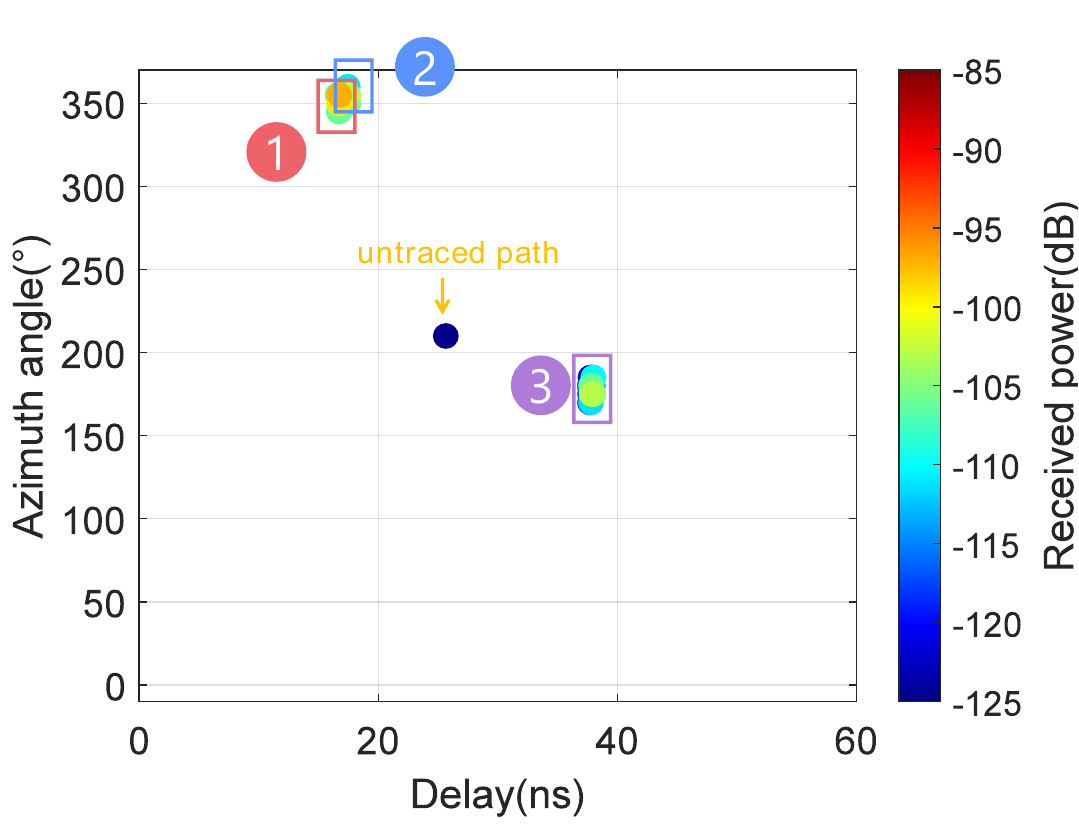}
    }
	\hfill
	\subfloat[]{
		\includegraphics[width=0.77\linewidth]{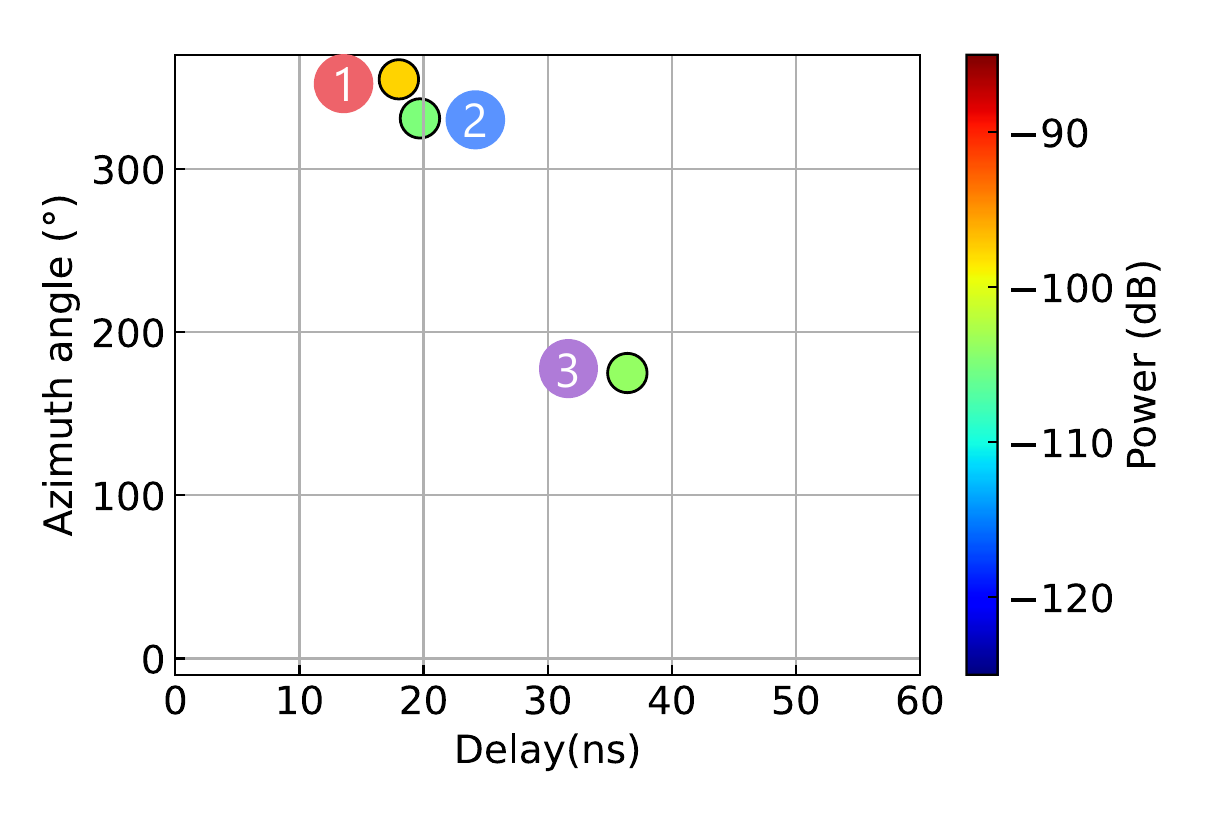}
	}
	\caption{PADP of (a) Physical Twin; (b) Calibrated RT Twin.}
	\label{fig:PADP}
\end{figure}

Fig.~\ref{fig:PADP} compares the corresponding PADPs obtained from the measurement and the calibrated RT simulation. It can be observed that the calibrated RT successfully reconstructs all three dominant propagation paths with high accuracy. Specifically, the delay errors of the three dominant MPCs are only 0.072~ns, 0.350~ns, and 0.241~ns, respectively, while the corresponding power differences are within 1~dB, demonstrating the effectiveness of the geometry and material calibration.
Nevertheless, an obvious discrepancy can still be observed around the second cluster. As indicated by the untraced path in Fig.~\ref{fig:PADP}(a), the measurement contains an additional weak MPC that cannot be reproduced by the deterministic RT simulation. This path is mainly generated by diffuse scattering from irregular rack surfaces and surrounding equipment, which are not explicitly modeled by conventional specular ray tracing. Consequently, the calibrated RT captures the dominant propagation mechanisms but still underestimates the diffuse multipath components observed in practical environments.

The influence of these missing diffuse components can be further quantified through the RMS delay spread. The measured RMS delay spread is 0.3923~ns, whereas the deterministic RT predicts only 0.3691~ns, corresponding to an underestimation of approximately 5.9\%. After incorporating the proposed hybrid propagation model, the predicted RMS delay spread increases to 0.3957~ns, which differs from the measurement by less than 1\%. This result demonstrates that the hybrid model effectively compensates for the insufficient diffuse scattering representation of conventional RT while preserving the dominant propagation characteristics.
Overall, the above observations indicate that the calibrated RT twin accurately reconstructs the dominant LoS and specular reflection paths, while the proposed hybrid propagation model further recovers the missing diffuse scattering components. Together, they provide a high-fidelity physical twin that serves as the foundation for the subsequent AI channel twin and system-level DT.

\subsection{AI Twin}
\begin{figure*}
    \centering
    \includegraphics[width=0.9\linewidth]{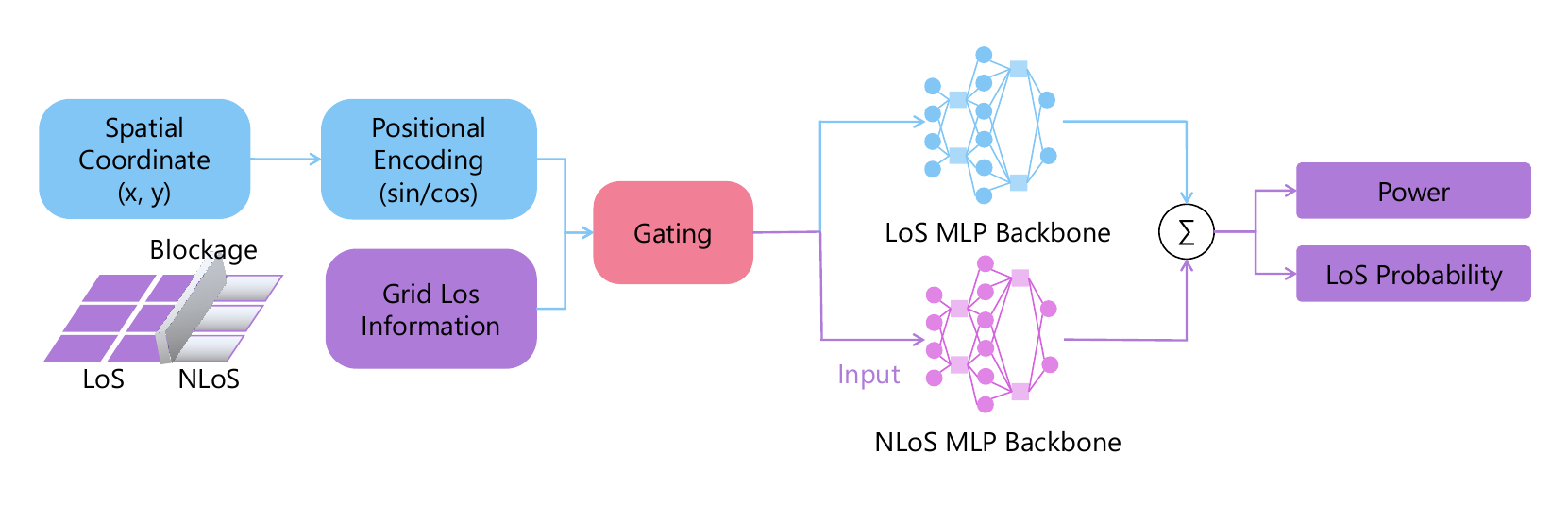}
    \caption{RT-conditioned implicit neural field.}
    \label{fig:inf}
\end{figure*}

While the measurement-calibrated RT twin provides physically interpretable channel predictions, its computational complexity remains a major challenge for real-time network evaluation and optimization. 
For example, the data center scenario requires computation over thousands of wireless links~\cite{11605106}, making full RT simulations computationally prohibitive.
To address this limitation, an AI twin is incorporated into the proposed DT framework as an acceleration engine. 


To capture the blockage-sensitive propagation characteristics of THz wireless data centers, a LoS-aware INF architecture in Fig.~\ref{fig:inf} is developed. The network takes the transmitter-receiver geometry as input, including spatial coordinates, propagation distance, and a grid-based LoS probability prior derived from the geometry twin. Positional encoding is further employed to enhance the representation capability of high-frequency spatial variations.
Unlike conventional coordinate-based neural fields that utilize a single regression network~\cite{VTCzmj}, the proposed architecture incorporates a LoS-adaptive dual-expert mechanism. Specifically, a shared encoder first extracts common spatial features, which are subsequently processed by dedicated LoS and NLoS expert branches. A gating module then adaptively combines the outputs of the two experts according to the local visibility condition. Such a design explicitly reflects the fundamentally different propagation behaviors of LoS and NLoS links in blockage-dominated THz environments, where path-loss characteristics, shadowing effects, and multipath contributions exhibit significantly different statistical properties.

\begin{algorithm}[t]
\caption{Construction of the AI-Assisted Channel Twin}
\label{alg:aitwin}
\begin{algorithmic}[1]

\Require
Calibrated RT environment $\mathcal{E}$,
receiver region $\mathcal{A}$,
grid size $\Delta$,
number of samples per grid $N_s$

\Ensure
Trained AI channel twin $\mathcal{F}_{\theta}$

\State Divide $\mathcal{A}$ into multiple grids ${\mathcal{G}_m}$

\For{each grid $\mathcal{G}_m$}

\State Randomly generate $N_s$ receiver locations

\For{each sampled receiver position}

    \State Run RT simulation

    \State Extract channel parameters
    $\{P_r,\tau,\phi,\theta\}$

    \State Determine LoS/NLoS condition

\EndFor

\State Compute LoS probability
$p_{L,m}$
\EndFor

\State Construct training dataset
$\mathcal{D}$

\For{each sample in $\mathcal{D}$}

\State Generate positional encoding

\State Query corresponding grid-based LoS probability

\State Form input feature
$[x,y,d,p_L]$

\EndFor

\State Train LoS-adaptive dual-expert neural field

\State Obtain AI channel twin
$\mathcal{F}_{\theta}$

\end{algorithmic}
\end{algorithm}

The proposed LoS-aware INF model explicitly incorporates the distance-dependent propagation prior and the grid-based LoS probability map obtained from the geometry twin.
Let $\mathbf u=[x,y]^\top$ denote the 2D receiver coordinate, and let $d$ be the Tx-Rx distance. The neural field is formulated as a nonlinear mapping function
\begin{equation}
\mathcal{F}_{\theta}:(\mathbf{u}, d, p_L)\rightarrow\mathbf{y},
\end{equation}
where $\mathbf{u}=(x,y)$ denotes the receiver location, $d$ is the Tx-Rx distance, $p_L$ is the grid-based LoS prior, and $\mathbf{y}$ represents the predicted channel response, including received power and optionally delay/angular statistics.

To improve numerical stability and enable multi-scale representation learning, the input features are normalized as
\begin{equation}
\tilde{\mathbf{u}} =\frac{\mathbf{u}-\mu_u}{\sigma_u}, \quad\tilde{d} =\frac{d-\mu_d}{\sigma_d}.
\end{equation}

Furthermore, positional encoding is applied to capture high-frequency spatial variations in THz propagation:

\begin{equation}
\gamma(\mathbf{x})=\left[\sin(2^k \pi \mathbf{x}),\cos(2^k \pi \mathbf{x})\right]_{k=0}^{K}.
\end{equation}






The model is trained by jointly optimizing the power prediction error and the LoS gating error. The power regression loss is defined as
\begin{equation}
\mathcal L_{\mathrm{power}}
=
\frac{1}{N}
\sum_{n=1}^{N}
\left(
\hat P^{(n)} - P^{(n)}_{\mathrm{gt}}
\right)^2,
\end{equation}
while the LoS classification loss is given by the binary cross-entropy,
\begin{equation}
\mathcal L_{\mathrm{LoS}}
=
-\frac{1}{N}
\sum_{n=1}^{N}
\left[
y^{(n)}_{\mathrm{LoS}}\log \hat p_L^{(n)}
+
\left(1-y^{(n)}_{\mathrm{LoS}}\right)\log\left(1-\hat p_L^{(n)}\right)
\right],
\end{equation}
where $y_{\mathrm{LoS}}\in\{0,1\}$ is the ground-truth LoS label obtained from the grid-based visibility map. The overall training objective is
\begin{equation}
\mathcal L
=
\lambda_1 \mathcal L_{\mathrm{power}}
+
\lambda_2 \mathcal L_{\mathrm{LoS}},
\end{equation}
where $\lambda_1$ and $\lambda_2$ are weighting factors used to balance power reconstruction and LoS-aware propagation learning.

\begin{figure}
    \centering
    \includegraphics[width=0.8 \linewidth]{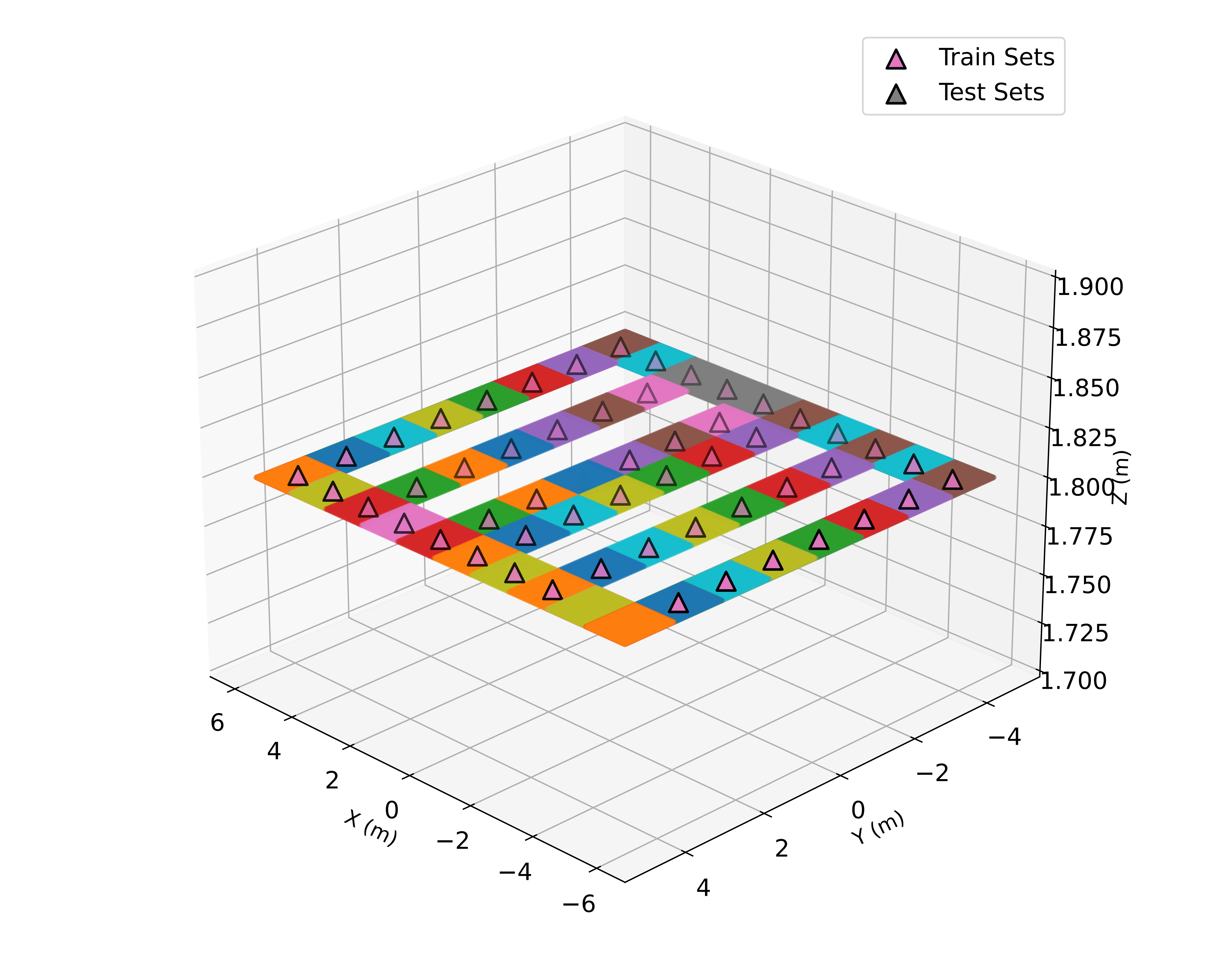}
    \caption{Training sets and test sets for AI Twin.}
    \label{fig: sets}
\end{figure}

To construct the dataset for AI twin training, the measurement plane is first partitioned into multiple spatial regions, as illustrated in Fig.~\ref{fig: sets}. A region-based spatial partition strategy is adopted to better evaluate the generalization capability of the AI twin.
Specifically, the entire data-center area is divided into a series of non-overlapping sub-regions. The colored regions are selected as training sets, while the white regions are reserved as test sets. For each training region, a large number of receiver locations are generated through spatial sampling, and the corresponding channel parameters are obtained from the calibrated RT twin. The generated channel database includes the receiver coordinates, propagation distance, LoS probability, and channel characteristics such as received power, delay, and angular information.
It is noteworthy that the testing regions are completely excluded from the training process. Such a spatial hold-out strategy provides a more rigorous evaluation of the model's capability to learn the underlying propagation mechanisms and reconstruct continuous channel fields in blockage-dominated THz data-center environments.

\begin{figure}
    \centering
    \includegraphics[width=0.8\linewidth]{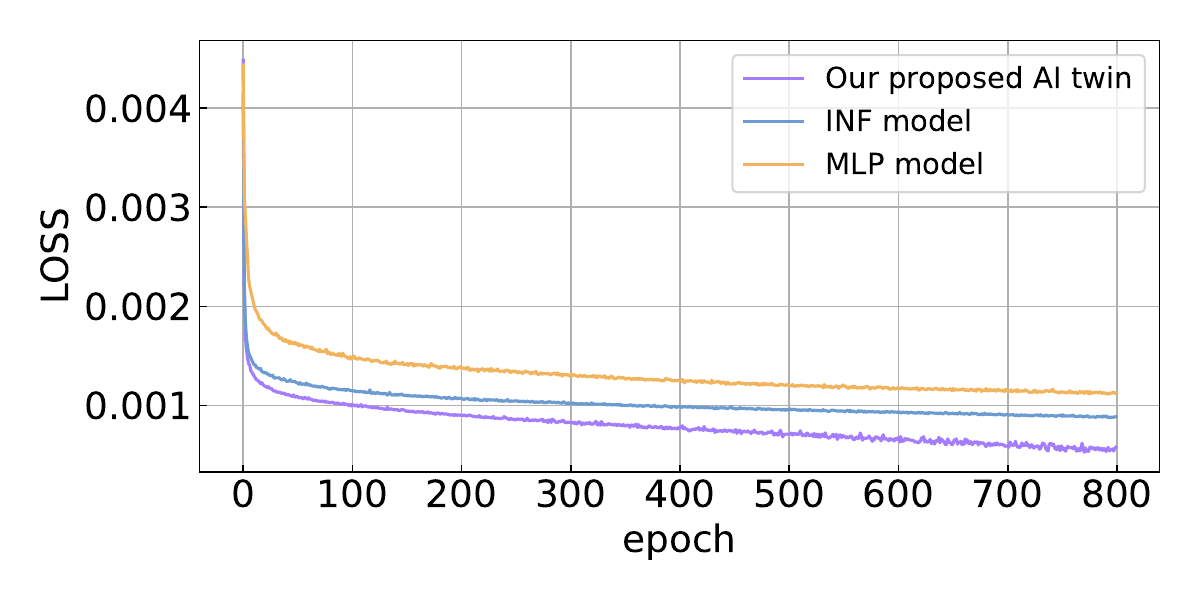}
    \caption{Training loss of different AI model.}
    \label{fig:loss}
\end{figure}
Fig.~\ref{fig:loss} compares the training convergence of the proposed AI channel twin, the conventional INF-based model, and the simple MLP baseline.
It can be observed that all models converge rapidly during the initial training stage, indicating that the large-scale path-loss trend can be efficiently learned from the generated channel dataset. However, the proposed model consistently achieves the lowest training loss throughout the entire optimization process. After 800 epochs, the loss of the proposed model decreases to approximately $5\times10^{-4}$, significantly outperforming both the INF-based and MLP-based baselines.
The faster convergence and lower final loss indicate that the proposed architecture provides a more physically consistent representation of THz propagation than generic coordinate-based neural fields.

\begin{figure*}
    \centering
    \subfloat[]{
        \includegraphics[width=0.31\linewidth]{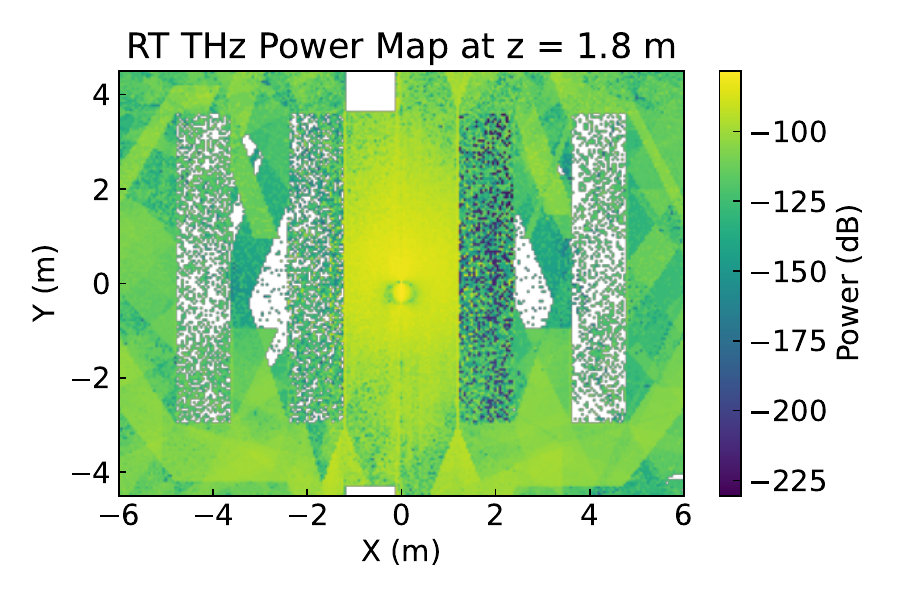}
    }
    \subfloat[]{
    \includegraphics[width=0.31\linewidth]{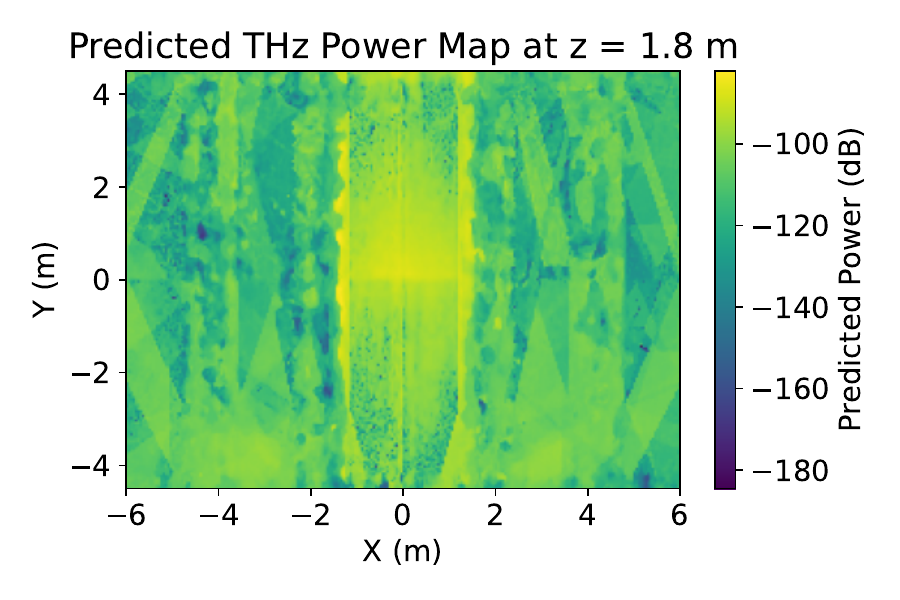}
    }
    \subfloat[]{
    \includegraphics[width=0.31\linewidth]{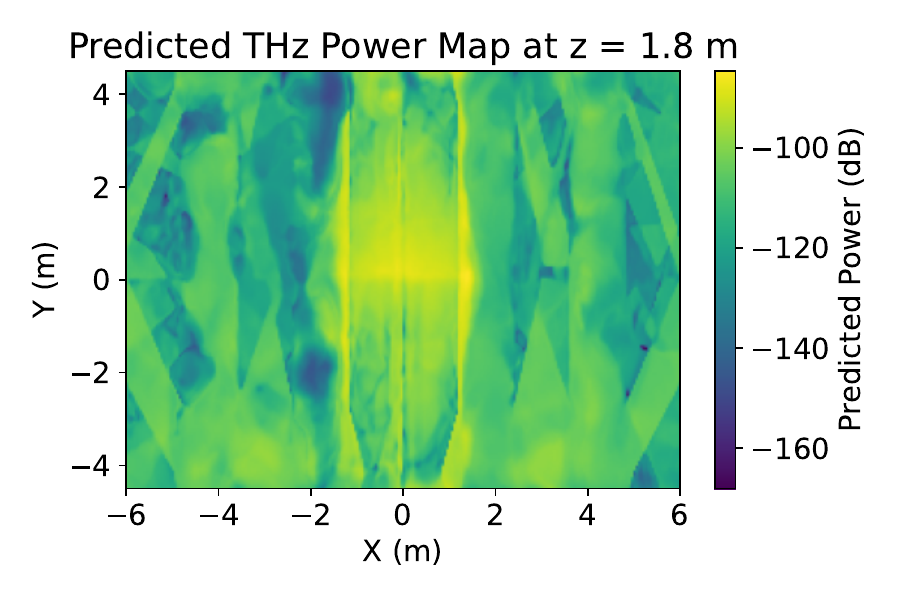}
    }
    \hfill
    \subfloat[]{
        \includegraphics[width=0.31\linewidth]{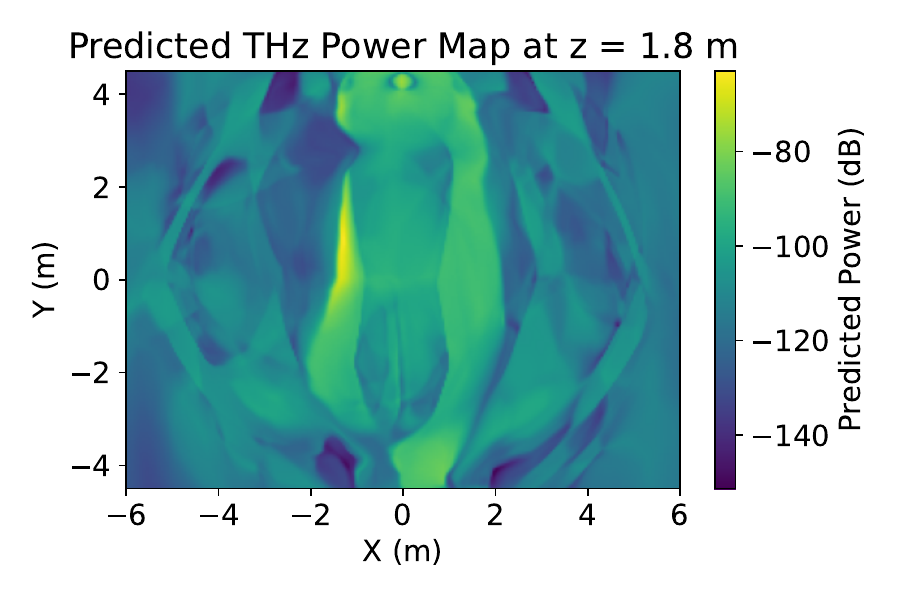}
    }
    \subfloat[]{
        \includegraphics[width=0.31\linewidth]{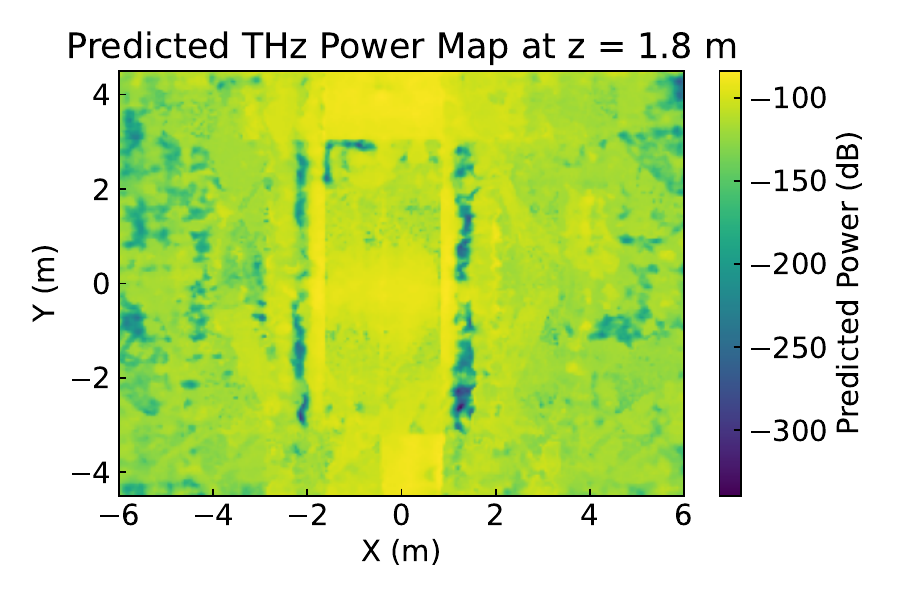}
    }
    \subfloat[]{
        \includegraphics[width=0.31\linewidth]{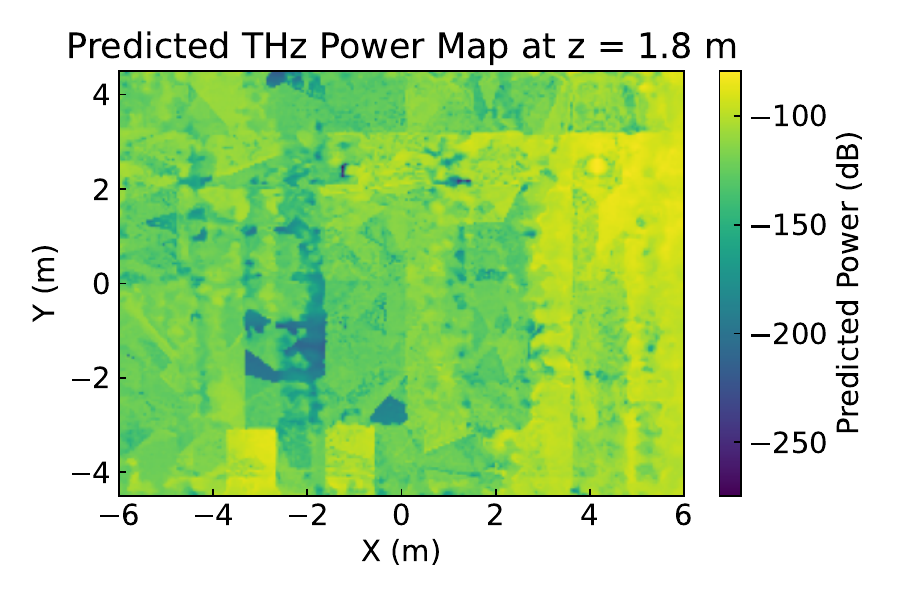}
    }
    \caption{Received power map. (a) RT Twin; (b) Our proposed AI twin; (c) INF-based AI Twin; (d) Simple MLP-based AI Twin; (e) Proposed AI twin (corridor Tx case); (f) Proposed AI twin (Corner Tx case)}
    \label{fig:field}
\end{figure*}

Fig.~\ref{fig:field} compares the reconstructed power maps generated by different AI channel twins against the calibrated RT ground truth.
As shown in Fig.~\ref{fig:field}(a), the calibrated RT twin reveals strong spatial variations caused by rack blockage, corridor-guided propagation, and specular reflections. Distinct LoS corridors can be observed along the central aisle, while deep shadowing regions are formed behind the rack structures.
Fig.~\ref{fig:field}(b) presents the reconstruction results of the proposed AI channel twin. The dominant propagation structures are accurately reproduced, including the blockage boundaries, reflected regions, and corridor-enhanced propagation zones. More importantly, the reconstructed field exhibits smooth spatial transitions while preserving the sharp power discontinuities introduced by rack blockage, demonstrating that the proposed model successfully learns both large-scale path-loss trends and blockage-dependent propagation mechanisms.
In contrast, the conventional INF model shown in Fig.~\ref{fig:field}(c) suffers from noticeable over-smoothing. Although the general propagation trend can still be reconstructed, the blockage boundaries become significantly blurred, and several deep-shadow regions disappear. This observation suggests that positional encoding alone is insufficient for representing the highly discontinuous propagation characteristics of blockage-dominated THz environments.

The simple MLP baseline shown in Fig.~\ref{fig:field}(d) exhibits even larger reconstruction errors. Due to the absence of propagation-aware priors, the predicted power field contains unrealistic spatial artifacts and fails to preserve the underlying physical propagation structures. Consequently, the reconstructed field deviates substantially from the RT-generated reference.
To further evaluate the generalization capability of the proposed AI twin, two additional transmitter configurations are considered, including a corridor-centered deployment and a corner deployment, as illustrated in Fig.~\ref{fig:field}(e) and Fig.~\ref{fig:field}(f).
Despite the significant change in the transmitter position, the proposed model remains capable of reconstructing the dominant propagation structures and blockage patterns. This result demonstrates that the proposed architecture can learn the underlying propagation mechanisms of the calibrated DT.

\begin{table}[]
    \caption{Performance Comparison of Different AI Channel Twins.}
    \label{tab:runtime}
    \centering
    \begin{tabular}{c|ccc}
       Model  & RMSE in dB & Runtime (ms)  \\
       \toprule
       MLP  & 7.504 &  138.7\\
       INF & 6.302 &  75 \\
       Proposed & 5.881 &  143 \\
       RT & - & $1.8 \times 10^6 $ \\
    \end{tabular}
\end{table}


Table~\ref{tab:runtime} compares the reconstruction accuracy and inference latency of different channel reconstruction approaches. Compared with the conventional MLP and INF models, the proposed LoS-aware AI twin achieves the lowest reconstruction error while maintaining real-time inference capability. Although the introduction of the grid-based LoS prior and dual-expert architecture slightly increases the inference time, the overall latency remains within hundreds of milliseconds, which is negligible for offline digital-twin construction and network evaluation. 
The RT results are not included in the table because they serve as the reference on the test set.
More importantly, all learning-based approaches significantly outperform conventional RT in terms of computational efficiency. While a complete RT simulation requires approximately $1.8\times10^6$~ms for dense channel reconstruction, the proposed AI twin completes the same task within only 143~ms, corresponding to more than four orders of magnitude acceleration. Such computational efficiency enables rapid channel prediction over large deployment areas and provides practical support for real-time coverage evaluation and deployment optimization.
Overall, the proposed AI twin achieves a favorable trade-off between reconstruction accuracy and computational complexity. By incorporating propagation-aware priors into the neural field, the proposed model preserves the dominant physical characteristics of THz propagation while maintaining sufficiently low inference latency for system-level DT applications.



In conclusion, the proposed AI channel twin is capable of efficiently reconstructing spatially continuous channel maps, including received power, path-loss distributions, and LoS probability fields, while preserving the dominant propagation mechanisms observed in the calibrated RT simulations.
By coupling the high-fidelity RT twin with the lightweight AI twin, the proposed framework achieves a favorable trade-off between physical accuracy and computational efficiency, forming the foundation for real-time digital-twin-assisted optimization in future THz wireless data centers.

\section{Evaluation layer}
\label{section: evaluation}

The evaluation layer translates channel-level information into network-level performance metrics empowered by the measurement layer. The evaluation layer bridges the gap between channel-level predictions and network-level performance assessment. Based on the outputs of the channel twin, including path loss, received power, delay dispersion, angular characteristics, and LoS probability, this layer evaluates the communication performance of the THz wireless data center through analytical and simulation-based approaches.

Specifically, the evaluation layer constructs spatially continuous maps of signal strength, interference, signal-to-interference-plus-noise ratio (SINR), coverage probability, and achievable rate throughout the data-center environment. Unlike conventional approaches that require repeated RT simulations for each deployment configuration, the proposed framework directly utilizes the outputs of the channel twin to estimate communication performance in real time.

Furthermore, analytical coverage and rate models are incorporated to establish a direct relationship between channel statistics and network performance indicators. By combining the predicted LoS probability, received-power statistics, and blockage information, closed-form coverage expressions can be derived, enabling rapid evaluation of network quality under different deployment strategies. Consequently, the evaluation layer provides a computationally efficient platform for large-scale performance assessment and forms the basis for subsequent optimization and decision-making procedures.

Based on the proposed DT, the AP deployment and resource allocation problem can be formulated as
\begin{subequations}
\label{opt_problem}
\begin{align}
    \max \limits_{\mathbf{V}} & \frac{1}{N_R} \sum_{i=1}^{N_R} \mathcal{R}(\mathbf{v}_0,~\mathbf{v}_i) + \sum_{0\leq i,j\leq N_R,i\neq j} \mathcal{R}(\mathbf{v}_i,\mathbf{v_j}) \label{subequatios:opt} \\ 
    \mathrm{s.t.} & ~ (x_i,y_i,z_i) \in V \label{subequation:V} \\ 
    &~ N_R \geq 1, ~N_P\in \mathbb{N} \label{subequation:N}\\
    &~ 1 \leq i,j \leq N_R ~\forall i,j \in \mathbb{N} \label{subequation:i} \\
    &~ P_c(\gamma_{i})>P_{th} \label{subequation:Pc}
\end{align}
\end{subequations}
where $\mathbf{V}$ denotes the set of AP deployment positions, $\mathcal{R}(\cdot)$ represents the achievable rate between communication nodes, and $P_c(\gamma_i)$ denotes the coverage probability under the SINR threshold $\gamma_i$.

To solve the optimization problem in (\ref{opt_problem}), the achievable rate and the corresponding SINR distribution should first be derived. Based on the constructed DT, system-level performance metrics such as interference and coverage~\cite{9247469} can be efficiently evaluated. The signal-to-interference-plus-noise ratio (SINR) at a receiver is expressed as
\begin{equation}
\begin{split}
    \mathrm{SINR} = \frac{S}{I + P_N} = \frac{P_tG_0K_ug(u)}{\sum_{i\in{\Phi_0/AP_0}}P_tG_iK_{x_i}g(x_{i}) + N_0B},
\end{split}
\label{equation: SINR}
\end{equation}
where $P_t$ is the transmit power, $G_0$ and $G_i$ denote the antenna gains of the desired and interfering links, $K_u$ and $K_{x_i}$ capture path loss and shadowing effects obtained from the DT, $g(\cdot)$ represents small-scale fading, $N_0$ is the noise spectral density, and $B$ is the system bandwidth.
Coverage probability is defined as the probability that the received SINR exceeds a predefined threshold $T$, i.e.,
\begin{equation} \begin{split} P_c(T)=\iiint_{V}(p_L(u)P_{c,L}(T)+p_N(u)P_{c,N}(T))f(v)\mathrm{d}V, \end{split} \label{equation: CP} \end{equation}
where $f(v)$ denotes the spatial probability density of receiver locations, while $p_L(u)$ and $p_N(u)$ denote the probabilities of LoS and NLoS propagation obtained from the proposed geometry twin.
The proposed analytical framework is applicable to two representative communication paradigms in wireless data centers, namely AP-to-rack communications and rack-to-rack communications.

\subsection{AP Deployment}
The AP-to-rack scenario corresponds to the infrastructure-based downlink transmission, where a ceiling-mounted AP communicates with multiple server racks. Owing to the highly directional antennas and severe propagation attenuation at THz frequencies, the interference from neighboring APs is typically much weaker than the thermal noise. Therefore, the communication system can be approximated as a noise-limited network,
\begin{equation}
	\mathrm{SINR}(u) \approx \frac{S(u)}{N_0 B}
\end{equation}
Following the measurement observations, the received power in the logarithmic domain is approximated by a Gaussian distribution,
\begin{equation}
S(u)\sim\mathcal N\left(\mu_S(u),\sigma_S^2(u)\right),
\end{equation}
where the location-dependent mean and variance are directly predicted by the proposed AI channel twin.
Consequently, the LoS conditional coverage probability can be derived in closed form as
\begin{equation}
	P_{c,L}(T) = Q\left(\frac{10\log_{10}(TN_0B)-\mu_{S}(u)}{\sigma_{S}(u)}\right)
\end{equation}
where $Q(\cdot)$ is the Gaussian Q-function. Similarly, the NLoS coverage probability is given by
\begin{equation}
	P_{c,N}(T) = Q\left(\frac{10\log_{10}(TN_0B)-\mu_{S,N}(u)}{\sigma_{S,N}(u)}\right)
\end{equation}
Therefore, the spatial coverage probability can be efficiently evaluated without invoking computationally expensive RT simulations. Instead, the AI twin predicts the local channel statistics, from which the coverage probability is directly obtained through the above closed-form expressions.
\begin{algorithm}[t]
\caption{Coverage Evaluation Using AI Channel Twin}
\label{alg:coverage}
\begin{algorithmic}[1]

\Require
AI channel twin $\mathcal{F}_{\theta}$,
coverage threshold $T$
\Ensure
Coverage probability $P_c(T)$
\For{each receiver location $u$}
\State Query grid-based LoS probability
\State Predict received power statistics
using $\mathcal{F}_{\theta}$
\State Compute conditional coverage
$P_{c,L}(T|u)$ and
$P_{c,N}(T|u)$
\EndFor
\State Evaluate
[
$P_c(T)=
\frac{1}{M}
\sum_{m=1}^{M}$
$\Big[
p_{L,m}P_{c,L}
+
(1-p_{L,m})P_{c,N}
\Big]$
]
\Return $P_c(T)$
\end{algorithmic}
\end{algorithm}

\begin{figure}
    \centering
    \subfloat[]{\includegraphics[width=0.85\linewidth]{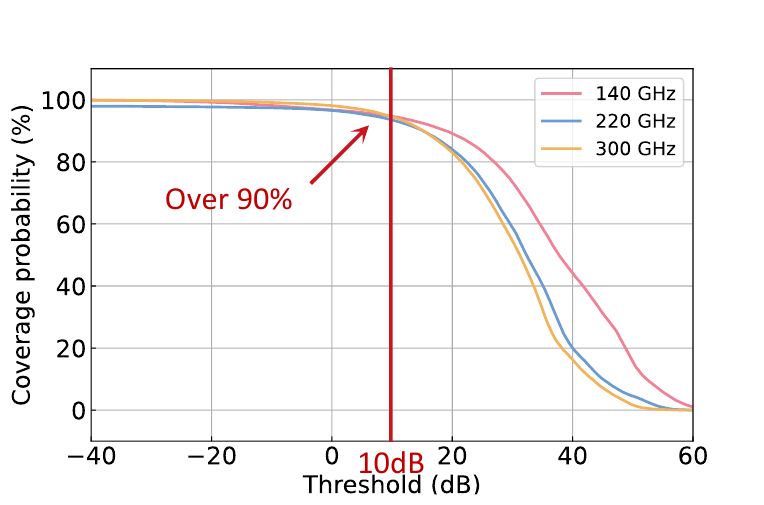}}
    \hfill
    \subfloat[]{\includegraphics[width=0.85\linewidth]{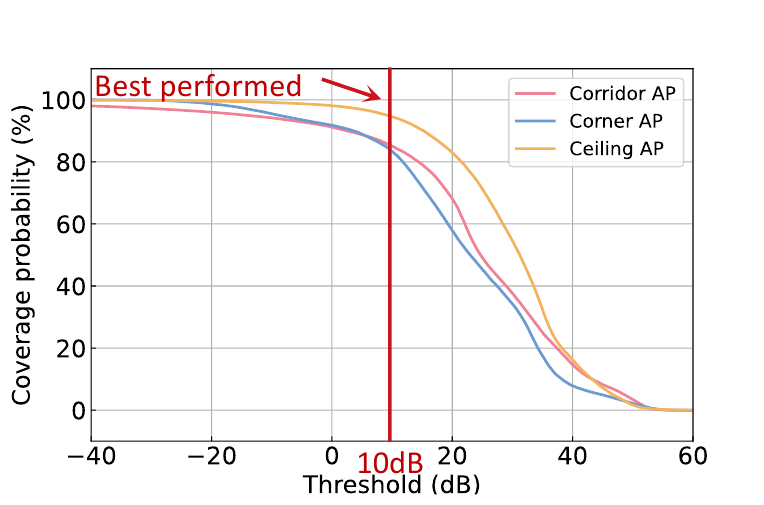}}
    \caption{Coverage Probability of (a) different frequency bands, and (b) different AP positions.}
    \label{fig: coverage}
\end{figure}

Fig.~\ref{fig: coverage}(a) presents the coverage probability under different carrier frequencies. As expected, the coverage probability monotonically decreases with the SINR threshold, since fewer receiver locations satisfy increasingly stringent communication requirements.
Among the three frequency bands, the 140~GHz system consistently achieves the highest coverage probability over the entire SINR range, followed by the 220~GHz and 300~GHz bands. The performance degradation at higher frequencies mainly results from the increased free-space path loss, stronger molecular absorption, and higher sensitivity to blockage, which jointly reduce the received signal power and consequently lower the achievable SINR. Although higher-frequency bands provide larger available bandwidth, their coverage becomes more limited under identical transmit power and antenna configurations. This observation demonstrates the inherent trade-off between spectrum availability and propagation robustness in THz wireless data centers.
It is worth noting that the coverage curves exhibit the steepest transition in the SINR range of approximately 15--35~dB. This indicates that the majority of receiver locations operate within this SINR region, where small improvements in channel quality can significantly increase the overall network coverage. Therefore, this operating region is particularly suitable for AP deployment optimization and beam management.

Fig.~\ref{fig: coverage}(b) compares the coverage performance under different AP deployment locations, including the corridor AP, corner AP, and ceiling-mounted AP configurations.
Among the three deployment strategies, the ceiling AP achieves the highest coverage probability over most SINR thresholds. Benefiting from its centrally elevated position, the ceiling deployment provides shorter average propagation distances and fewer blockage events, thereby maintaining stronger LoS connectivity to the majority of server racks.
The corridor AP exhibits intermediate performance. Although the transmitter is located inside the main propagation corridor, several racks remain partially shadowed by neighboring cabinets, leading to localized coverage degradation.
In contrast, the corner deployment consistently achieves the lowest coverage probability. Since most communication links experience longer propagation distances and multiple blockage events, both the received power and the LoS probability decrease significantly, resulting in inferior network coverage.

\subsection{Concurrent Rack-to-rack Communications}
Besides infrastructure-based access, future wireless data centers also require direct rack-to-rack communications, where multiple concurrent links coexist within the same deployment area. In this case, mutual interference among simultaneous transmissions cannot be neglected.
The channel gains between all rack pairs are first organized into a channel matrix,
\begin{equation}
    \mathbf{G}=\left[ \begin{array}{cccc}
        0 & g_{12} & g_{13} & \cdots \\
        g_{21} & 0 & g_{23} & \cdots \\
        g_{31} & g_{32} & 0 & \cdots \\
        \cdots & & &\\
    \end{array} \right],
\end{equation}
where $g_{ij}$ denotes the channel gain from rack $i$ to rack $j$ predicted by the DT.
Assuming equal transmit powers, the aggregate interference experienced by receiver $i$ is
\begin{equation}
I_i=\sum_{j\neq i}P_tg_{ji}b_j,
\end{equation}
where $b_j\in\{0,1\}$ denotes the transmission activity indicator.
Since the individual channel gains approximately follow lognormal distributions, the aggregate interference can be approximated by another lognormal random variable using the Fenton--Wilkinson approximation,
\begin{equation}
I_i\sim\ln\mathcal N(\mu_I,\sigma_I^2).
\end{equation}
The corresponding coverage probability is therefore
\begin{align}
P_c&=\mathbb{P} \left(\frac{S_i}{I_i+N}>T\right)\nonumber\\
&=\Phi\left(\frac{\frac{S_i}{T}-N-\mu_I}{\sigma_I}\right),
\label{eq:rack_coverage}
\end{align}
where $\Phi(\cdot)$ denotes the cumulative distribution function of the standard Gaussian distribution.
Compared with the AP-to-rack scenario, the rack-to-rack case explicitly captures the impact of concurrent transmissions and thus provides a more realistic performance evaluation for dense wireless data-center networks.
Therefore, the average rate can be derived as
\begin{equation}
    \bar{R}=\frac{B}{\ln2}\int_0^{\infty}\frac{P_c(T)}{1+T}\mathrm{d} T.
\end{equation}
Overall, the proposed analytical framework establishes the connection between the AI channel twin and the system-level DT. The channel twin estimates the location-dependent channel statistics, from which the achievable rate, interference, and coverage probability can be efficiently derived. Consequently, the evaluation layer bridges the gap between channel reconstruction and intelligent network optimization, providing the theoretical foundation for the subsequent manipulation layer.

\begin{figure}
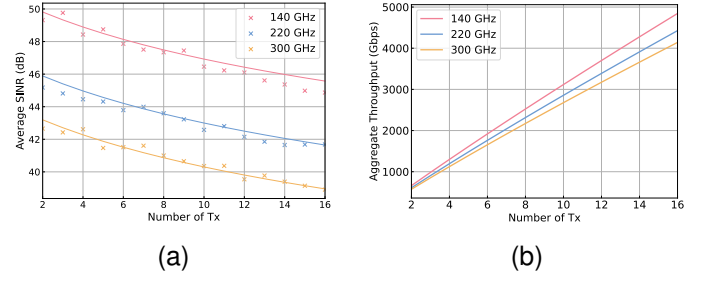

    \centering
    \subfloat[]{
        \includesvg[width=0.5\linewidth]{figures/SINR6.svg}
    }
    \subfloat[]{
        \includesvg[width=0.5\linewidth]{figures/Throughput.svg}
    }
    \caption{(a) Average SINR and (b) Aggregate throughput of concurrent rack-to-rack communications}
    \label{fig:r2r}
\end{figure}

Fig.~\ref{fig:r2r} presents the average SINR of the rack-to-rack communication scenario as the number of simultaneously active transmission links increases. The analytical results are compared with Monte Carlo simulations for the three measured frequency bands.
As expected, the average SINR gradually decreases with the number of active communication links due to the accumulation of co-channel interference. When only a few links are simultaneously active, the interference level remains relatively low, and the network is primarily limited by thermal noise. As more rack pairs are activated, the aggregate interference becomes the dominant performance bottleneck, resulting in a continuous degradation of the received SINR.
Despite the SINR degradation, the aggregate throughput increases with the number of concurrent links. This indicates that the throughput gain from spatially concurrent transmissions outweighs the loss in per-link spectral efficiency within the considered deployment range. Among the three frequency bands with the same bandwidth of $20$~GHz, the 140-GHz system achieves the highest aggregate throughput, followed by the 220- and 300-GHz systems.






\section{Conclusion}
\label{section: con}
We have proposed a measurement-driven multi-layer DT framework for THz wireless data centers. In the measurement layer, tri-band measurements at 140, 220, and 300 GHz are first conducted to capture the propagation characteristics of a realistic data-center environment. In the construction layer, a calibrated RT twin is constructed by jointly optimizing the geometry, material, antenna, and propagation models. An AI twin is then developed to enable efficient channel reconstruction with high accuracy and low latency. In the evaluation layer, the reconstructed channel field is used to build an evaluation twin for coverage and interference analysis. In the manipulation layer, power allocation and beam manipulation technologies~\cite{11267239}, e.g., airy beam design, can be utilized to complete the closed-loop optimization.
The proposed framework establishes an end-to-end pipeline from channel measurement to network optimization, bridging physical fidelity and computational efficiency. The experimental results validate the effectiveness of the measurement-calibrated RT twin and demonstrate that the proposed AI twin can accurately reconstruct the channel field while significantly accelerating inference. In addition, the system-level results confirm the potential of the proposed DT for THz wireless data-center planning and deployment optimization, indicating that the centered AP below the ceiling can provide the best coverage for the wireless data center.

\bibliographystyle{IEEEtran}
\bibliography{DC}

\newpage

\vfill

\end{document}